\documentclass[letterpaper]{article} % DO NOT CHANGE THIS
\usepackage{aaai25}  % DO NOT CHANGE THIS
\usepackage{times}  % DO NOT CHANGE THIS
\usepackage{helvet}  % DO NOT CHANGE THIS
\usepackage{courier}  % DO NOT CHANGE THIS
\usepackage[hyphens]{url}  % DO NOT CHANGE THIS
\usepackage{graphicx} % DO NOT CHANGE THIS
\usepackage{natbib}  % DO NOT CHANGE THIS AND DO NOT ADD ANY OPTIONS TO IT
\usepackage{caption} % DO NOT CHANGE THIS AND DO NOT ADD ANY OPTIONS TO IT
\usepackage{algorithm}
\usepackage{algorithmic}
\usepackage{amsmath}
\usepackage{booktabs}
\usepackage{multirow}
\usepackage{tabularx}
\usepackage{xcolor}
\usepackage{amsmath}
\usepackage{amssymb}   % 用于 \checkmark 符号
\usepackage{graphicx}  % 用于 \resizebox

\usepackage{newfloat}
\usepackage{listings}
\DeclareCaptionStyle{ruled}{labelfont=normalfont,labelsep=colon,strut=off} % DO NOT CHANGE THIS
\floatstyle{ruled}
\newfloat{listing}{tb}{lst}{}
\floatname{listing}{Listing}
\newcommand{\mymethod}[1]{Scene$\mathcal{X}$}
\usepackage{bibentry}

\title{\mymethod{}: Procedural Controllable Large-scale Scene Generation}

\author{
   Mengqi Zhou\textsuperscript{\rm 1,3,4,6}\textsuperscript{*},
   Yuxi Wang\textsuperscript{\rm 2}\textsuperscript{*},
   Jun Hou\textsuperscript{\rm 2},
   Shougao Zhang\textsuperscript{\rm 8},\\
   Yiwei Li\textsuperscript{\rm 1},
   Chuanchen Luo\textsuperscript{\rm 5},
   Junran Peng\textsuperscript{\rm 7}\textsuperscript{$\dag$},
   Zhaoxiang Zhang\textsuperscript{\rm 1,2,3,4,6}
}

\usepackage{bibentry}
\begin{document}

\twocolumn[{%
\renewcommand\twocolumn[1][]{#1}
\maketitle

\vspace{-2cm}
\begin{center}
\small
$^1$University of Chinese Academy of Sciences (UCAS),
$^2$Centre for Artificial Intelligence and Robotics (CAIR)\\
$^3$Institute of Automation, Chinese Academy of Sciences (CASIA),
$^4$New Laboratory of Pattern Recognition (NLPR)\\
$^5$Shandong University,
$^6$State Key Laboratory of Multimodal Artificial Intelligence Systems (MAIS)\\
$^7$University of Science and Technology Beijing,
$^8$China University of Geosciences Beijing\\
\texttt{{\{zhoumengqi2022, zhaoxiang.zhang\}@ia.ac.cn,yuxiwang93@gmail.com,jrpeng4ever@126.com}}

\end{center}

% \vspace{-0.2cm}  % 添加额外的间距调整
\begin{center}
    \centering
    \captionsetup{type=figure}
    \includegraphics[width=1\textwidth]{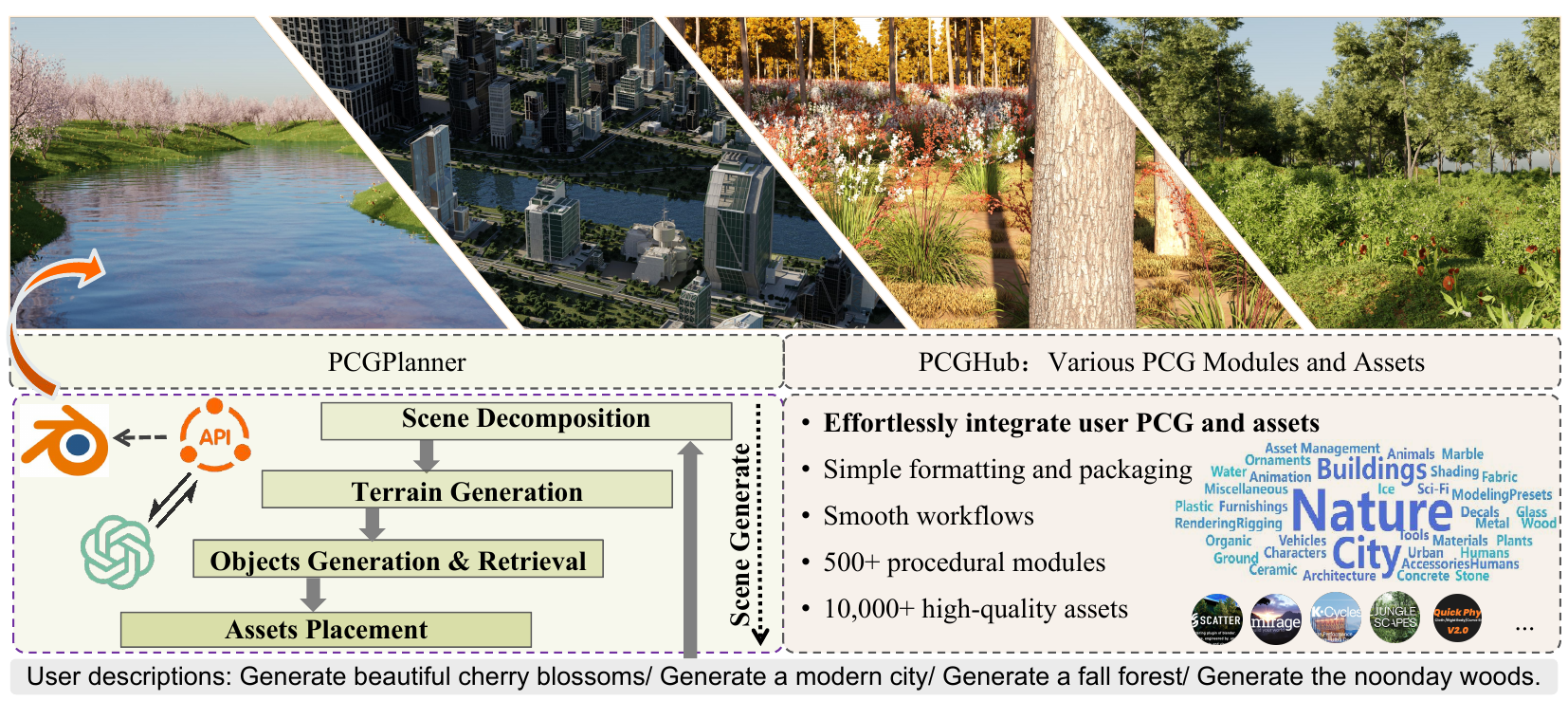}
    \vspace{-0.6cm}
    \caption{The proposed \mymethod{} can create large-scale 3D natural scenes or unbounded cities automatically according to user instructions. The generated models are characterized by delicate geometric structures, realistic material textures, and natural lighting, allowing for seamless deployment in the industrial pipeline.}
    \label{fig:Fig1}
    \vspace{0.4cm}
\end{center}
}]

% \twocolumn[{%
% \renewcommand\twocolumn[1][]{#1}%
% \maketitle
% \begin{center}
%     \centering
%     \captionsetup{type=figure}
%     \includegraphics[width=1\textwidth]{figs/SceneX_fig1.pdf} 
%     \captionof{figure}{The proposed \mymethod{} can create large-scale 3D natural scenes or unbounded cities automatically according to user instructions. The generated models are characterized by delicate geometric structures, realistic material textures, and natural lighting, allowing for seamless deployment in the industrial pipeline.}
%     \label{fig:Fig1}
% \end{center}
% }]

% \footnote{$^\ast$Equal contribution\quad} \\
% \footnote{$\dag$Corresponding author}
\renewcommand{\thefootnote}{}  % 完全不显示数字
\footnotetext{\textsuperscript{*}Equal contribution}
\renewcommand{\thefootnote}{\textsuperscript{$\dag$}}
\footnotetext{Corresponding author}

% \footnote{$^\ast$Equal contribution \quad \\
% $^\dagger$Corresponding author
% }
% \footnote{$^\dagger$Corresponding author}    

\begin{abstract}
Developing comprehensive explicit world models is crucial for understanding and simulating real-world scenarios. Recently, Procedural Controllable Generation (PCG) has gained significant attention in large-scale scene generation by enabling the creation of scalable, high-quality assets. However, PCG faces challenges such as limited modular diversity, high expertise requirements, and challenges in managing the diverse elements and structures in complex scenes. In this paper, we introduce a large-scale scene generation framework, \mymethod{}, which can automatically produce high-quality procedural models according to designers' textual descriptions. Specifically, the proposed method comprises two components, PCGHub and PCGPlanner. The former encompasses an extensive collection of accessible procedural assets and thousands of hand-craft API documents to perform as a standard protocol for PCG controller. The latter aims to generate executable actions for Blender to produce controllable and precise 3D assets guided by the user's instructions. Extensive experiments demonstrated the capability of our method in controllable large-scale scene generation, including nature scenes and unbounded cities, as well as scene editing such as asset placement and season translation. Our project page: https://zhouzq1.github.io/SceneX/.
\end{abstract}


\section{Introduction}
In the realm of Artificial General Intelligence (AGI), developing comprehensive world models is crucial for understanding and simulating real-world scenarios\cite{OOD-HOI, Stablemofusion}. Recent advancements, such as those demonstrated by the Sora model~\cite{sora}, show progress in capturing physical laws and generating realistic simulations. However, Sora's outputs often lack detailed geometry and structured information, limiting their editability and interactivity.
To address these limitations, explicit world models provide a more robust solution. By constructing worlds with detailed mesh assets and evolving them according to predefined physical rules, these models leverage Physically Based Rendering (PBR) to ensure high geometric consistency and detailed visualization.
Procedural modeling methods, such as those outlined by Lindenmayer et al. \cite{LINDENMAYER1968280}, show great promise in generating realistic and intricate world models through adjustable parameters and rule-based systems using tools like \textit{Blender}. For example, Infinigen \cite{Infinigen} proposes a procedural generator to generate large-scale natural scenes encompassing terrain, weather, vegetation, and wildlife. \cite{lipp2011interactive}, \cite{chen2008interactive} and \cite{talton2011metropolis} use procedural modeling to generate city-level street or layout. Although these procedural approaches generate high-quality 3D assets, they are beginner-unfriendly and time-consuming due to the comprehensive grasp of generation rules, algorithmic frameworks, and individual parameters of procedural modeling. %make the generation process beginner-unfriendly and time-consuming, especially for large-scale scene or city generation. 
% For instance, generating a New York city, as shown in Fig \ref{fig:Fig1}, requires the effort of a professional PCG engineer working for over two weeks. 
For instance, generating a city, as shown in Fig. \ref{fig:Fig1}, requires the effort of a professional PCG engineer working for over two weeks.%Moreover, due to the substantial learning curve for users, it is formidable to generate large-scale scenes using procedural modeling.

To address the above problems, existing works such as 3D-GPT \cite{3D-GPT} and SceneCraft \cite{hu2024scenecraft} introduce an instruction-driven 3D modeling method by integrating an LLM agent with procedural generation software. Despite the successful collaboration with human designers to establish a procedural generation framework, these methods still exhibit significant limitations. 
Firstly, the restricted editing capabilities of fixed 3D resources prevent SceneCraft from attaining the same degree of precision in 3D resource editing as is achievable through procedural generation. Secondly, 3D-GPT is based on the procedural generation model Infinigen, which restricts its capacity to fully utilise existing PCG resources. This is exemplified by the inability to generate extensive terrain and limitless cities. Finally, the task of generating a large-scale scene is often composed of multiple related subtasks, such as the planning of the scene layout, the generation and placement of assets, and the adjustment of environmental details. However, there is currently no effective way to manage the execution order and dependencies of these tasks.

In this paper, we introduce the \mymethod{} framework, consisting of PCGHub and PCGPlanner. 
%\textcolor{red}{PCGHub includes high-quality assets (both procedural and static), procedural placement systems (Scatter Generators), and procedural stylization components.} 
%PCGHub includes a vast array of accessible procedural assets and over a hundred handcrafted API files, serving as standard protocols for PCG modules.
PCGHub addresses the limitations of individual PCG modules, which are constrained by their inherent algorithms and predefined rules. By integrating diverse procedural modules and encapsulating them with corresponding APIs, PCGHub offers a platform for continuously incorporating new PCG capabilities, thus supporting the expansion and diversity of procedural generation resources.
Due to constraints in procedural algorithms and predefined rules, unrestricted combination of all procedural methods is infeasible. Thus, we develop the PCGPlanner to coordinate PCG methods within these constraints. PCGPlanner accesses and integrates procedural modules, enabling effective coordination and seamless integration. Our work improves scene generation flexibility and diversity, reduces entry barriers for non-coders, and efficiently uses existing technologies for accessible, community-friendly procedural generation. Our \mymethod{} possesses several key properties:
% \vspace{-0.5cm}
\begin{enumerate}
    \item \textit{Efficiency:} Benefiting from the proposed PCGHub and PCGPlanner, our \mymethod{} can rapidly produces extensive, high-quality 3D assets, including terrain, city, and forest. Moreover, we only need a few hours to generate a large-scale city, whereas it would take a professional designer over two weeks. 

    \item \textit{Controllability:} %Similar to text-to-3D generation \cite{Dreambooth3D}, 
    \mymethod{} can generate 3D scenes satisfying personalized demands. It achieves scene editing according to the corresponding descriptions, such as adding objects, placing objects at a location, translating the season, and so on. 

    \item \textit{Diversity:} \mymethod{} overcomes the constraints of conventional generation algorithms and predefined rules. By integrating and coordinating multiple subtasks, it attains a remarkable degree of flexibility and diversity in scene generation on a vast scale. 
    
\end{enumerate}

\section{Related Works}

\textbf{Learning Based 3D Generation.}
In recent years, 3D asset generation has witnessed rapid progress, combining the ideas of computer graphics and computer vision to realise the free creation of 3D content. Presently, predominant research in 3D asset generation primarily concentrate on creating individual objects~\cite{DreamFusion,Zero123,Magic3D,Fantasia3d,li2024materialseg3d,hu2024semantic}, 3D avatars~\cite{StyleaAvatar3d,DreamHuman,AvatarCLIP,AvatarCraft}, and 3D scenes \cite{Text2Room,Text2Nerf,SceneScape,SceneWiz3D,FurniScene}. Among these, ZeroShot123~\cite{Zero123} proposes a method based on a diffusion model, which realizes the 3D model of the target based on a picture. DreamFusion \cite{DreamFusion} proposes a NeRF-based approach that allows the model to generate a corresponding 3D model based on the input text. 
Compared with single object generation, it is more practical and challenging to generate large-scale scenes, including the generation of natural landforms~\cite{hao2021GANcraft,infinite_nature_2020,SceneDreamer} and borderless cities~\cite{InfiniCity,Matrixcity,xie2023citydreamer}.
CityDreamer~\cite{xie2023citydreamer} builds vast, large-scale 3D cities based on the layout of real cities, enhancing the accuracy and stability of urban reconstruction. SceneDreamer~\cite{SceneDreamer} proposes a method to generate 3D borderless scenes within 2D plots using BEV representations. %These learning-based methods do not have geometric structures, making them difficult to apply directly to virtual engines such as UE. For example, although InfiniCity \cite{InfiniCity} and CityDreamer \cite{xie2023citydreamer} can generate visually appealing urban scenes, the absence of clear geometry may lead to issues in practical applications, such as intersecting objects and spatial discontinuities.

\noindent\textbf{Procedural Based 3D Generation.}
Researchers have delved into the procedural generation of natural scenes \cite{ecosystems2022,rivers} and urban scenes~\cite{lipp2011interactive,talton2011metropolis,Parcels,10.1145/2508363.2508405} using Procedural Content Generation (PCG). For instance, PMC~\cite{PMC} proposes a procedural way to generate cities based on 2D ocean or city boundaries. It employs mathematical algorithms to generate blocks and streets and utilizes tailgating technology to generate the geometry of buildings. %However, these methods require specialized knowledge to fine-tune the algorithm. 
While these traditional computer graphic methods can produce high-quality 3D data, all parameters must be pre-entered into the procedurally generated process. %The resultant 3D data is bound by rule limitations, introducing a certain level of deviation from the real world. 
This significantly constrains flexibility and practical usability. Infinigen~\cite{Infinigen, cityX} introduces a technique for procedurally generating realistic 3D natural objects and scenes. %This methodology facilitates the programmatic generation of all assets, with meshes and textures generated through random mathematical algorithms. 
Although Infinigen generates infinitely assets, users are unable to customize the generated outcomes because of their specific requirements. In this paper, we propose a more convenient method to produce procedural assets.

\noindent\textbf{LLM Agents.}
Benefiting from knowledge hidden in the large-language model (LLMs) \cite{raffel2020exploring,achiam2023gpt,PaLM,BERT,bubeck2023sparks}, researchers explore LLMs to address intricate tasks beyond canonical language processing domains. These tasks encompass areas such as mathematical reasoning \cite{imani2023mathprompter,wei2022chain}, medicine \cite{yang2023evaluations,jeblick2023chatgpt}, and planning \cite{zhang2023proagent,gong2023mindagent,huang2023voxposer,huang2022language}. Thanks to powerful reasoning and generalization capabilities, LLMs act as practiced planners for different tasks. For example, \cite{huang2022language} utilizes the expansive domain knowledge of LLMs on the internet and their emerging zero-shot planning capabilities to execute intricate task planning and reasoning. \cite{gong2023mindagent} investigates the application of LLMs in scenarios involving multi-agent coordination, covering a range of diverse task objectives. \cite{zeng2022socratic} presents a modular framework that employs structured dialogue through prompts among multiple models. Moreover, specialized LLMs for particular applications have been explored, such as HuggingGPT \cite{shen2023hugginggpt} for vision perception tasks, VisualChatGPT \cite{wu2023visual} for multi-modality understanding, Voyager \cite{wang2023voyager} and \cite{zhu2023ghost}, SheetCopilot \cite{li2023sheetcopilot} for office software, and Codex \cite{Codex} for Python code generation. Inspired by existing works, we explore the LLM agent to PCG software, \textit{e.g., Blender,} to provide automatic 3D assets generation. 

\begin{figure}[!tbp]
\centering
\includegraphics[width=1\linewidth]{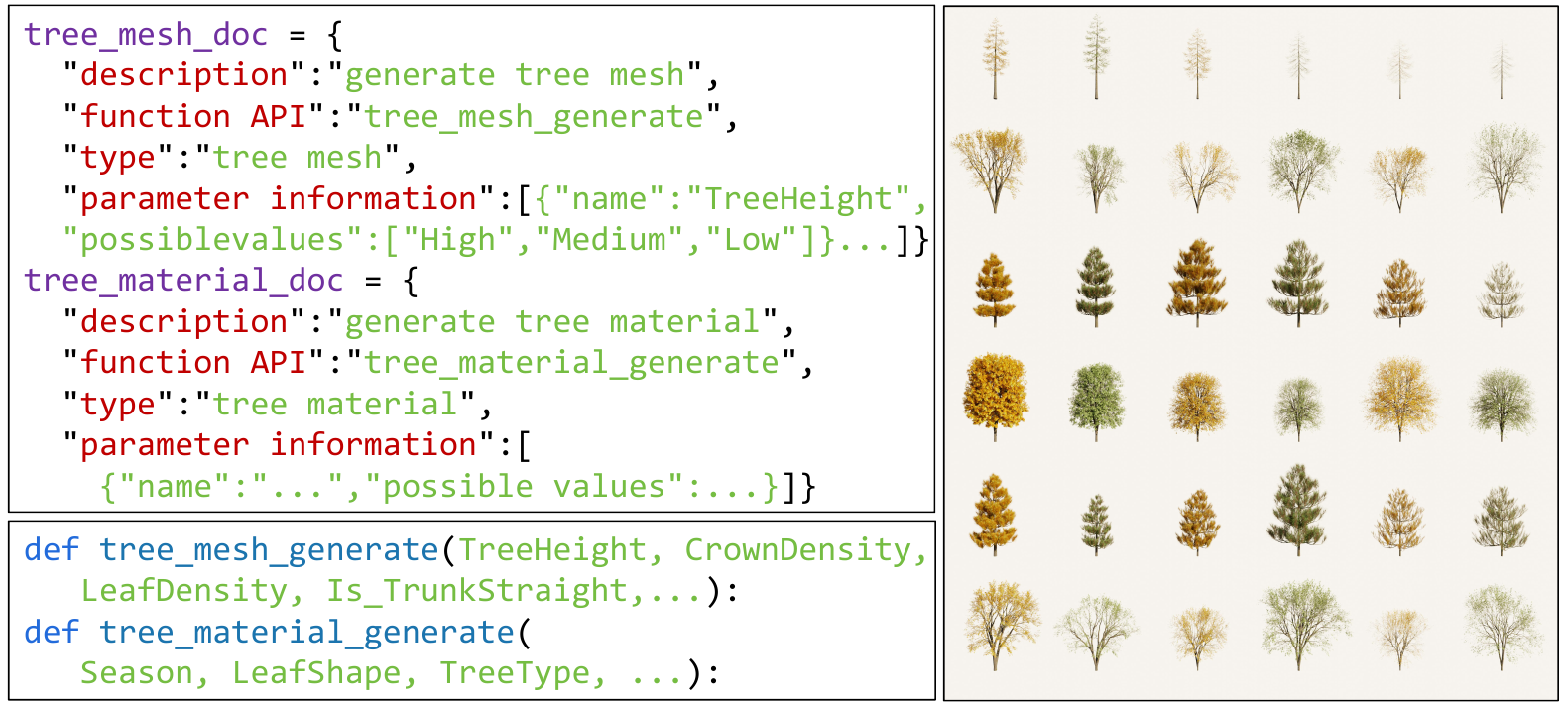}
% \caption{Illustration: Tree PCG API Documentation, API Functions, and Tree Generation Results}
\caption{Tree PCG API documentation, API functions, and tree generation results.}
\label{fig:tree}
\vspace{-0.45cm}
\end{figure}

\begin{table}[!tbp]
\centering
\footnotesize
%\vspace{-0.2cm} 
\resizebox{\columnwidth}{!}{%
\begin{tabular}{l|c|l|c} \toprule
 Capability & \ Num. PCG & Capability & \ Num. PCG \\
 \midrule
 Terrain & 12 & Water & 4 \\
 Weather & 15 & Snow & 3 \\
 Vegetation & 65 & Assets placement & 13 \\
 Buildings & 37 & Materials & 361 \\
 Blocks & 17 & Dynamic People & 4 \\
 Cities & 12 & Dynamic Vegetation & 23 \\
 People & 6 & Dynamic Vehicles & 3 \\
 \bottomrule
\end{tabular}%
}
\vspace{-0.2cm}
\caption{Overview of PCGHub capabilities.}
\label{tab:pcghub_capabilities}
\vspace{-0.76cm}
\end{table}

\section{\mymethod{}: All-in-One PCG Solution}
% \section{Methods}
%We introduce the \mymethod{} framework, which is designed for the versatile scene generation. \mymethod{} comprises two core components: PCGHub and PCGPlanner. PCGHub integrates a diverse range of 3D assets, including procedural and static types, alongside Scatter Generators and procedural stylization components. Due to the constraints of specific procedural algorithms and predefined rules, unrestricted combination of all procedural methods is not feasible. Therefore, we developed the PCGPlanner to coordinate procedural methods within these constraints. Procedural methods are wrapped with Python APIs and detailed documentation to facilitate their use by the PCGPlanner, enabling flexible integration and coordination of methods within the established limitations. Our framework lowers the entry barrier for users unfamiliar with coding or procedural generation, supports a variety of procedural methods, and efficiently leverages existing technologies to avoid redundancy, thereby ensuring both flexibility and variety in the generated scenes.
We present the \mymethod{} framework for versatile scene generation, which includes PCGHub and PCGPlanner. PCGHub integrates a vast array of procedural modules and 3D assets, offering extensive procedural capabilities, while PCGPlanner coordinates these procedural modules within a well-defined algorithmic structure. The framework seamlessly incorporates user-provided PCG and assets into the workflow. Leveraging user text input, the framework enables the efficient and precise generation of diverse, high-quality scenes.
%\textcolor{red}{With Python APIs and detailed documentation, our framework simplifies usage, supports various procedural methods, and efficiently leverages existing technologies, ensuring flexibility and variety in scene generation.}

\begin{figure}[!tbp]
\centering
\includegraphics[width=1\linewidth]{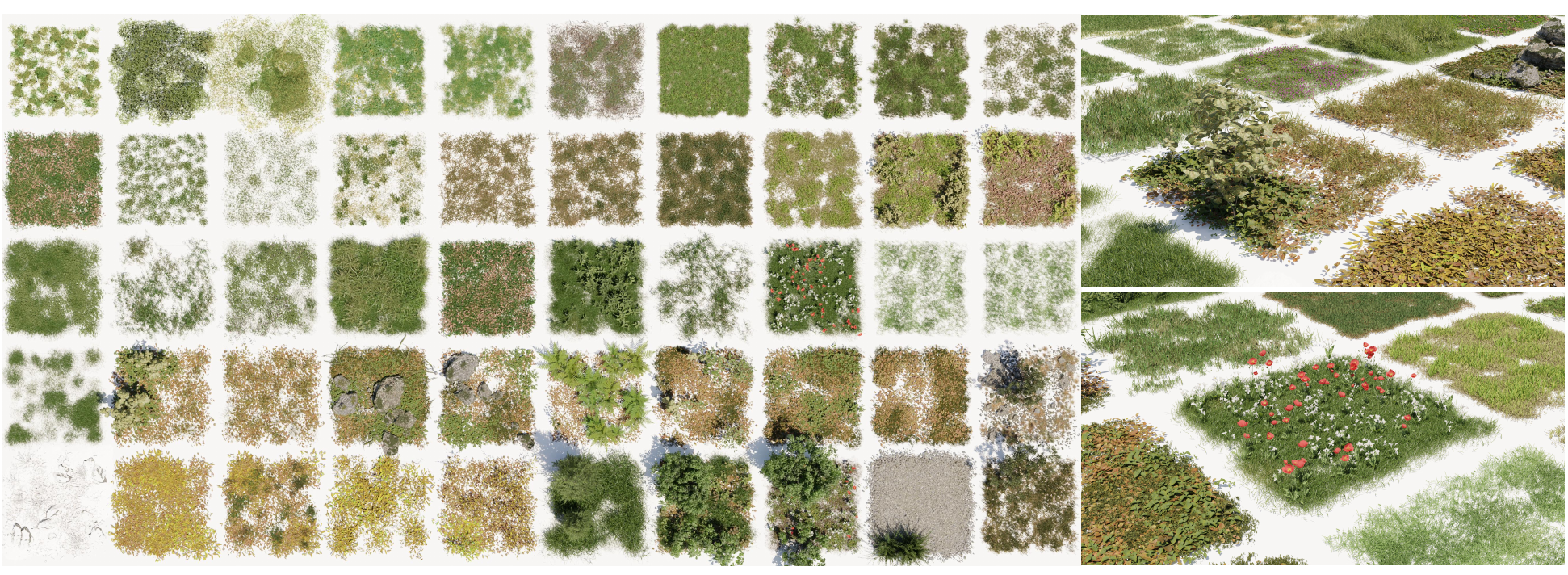}
%\vspace{-0.3cm}
\caption{Visualization of scatter layout results, showcasing diverse ground cover effects generated using various vegetation assets.}
\vspace{-0.2cm}
\label{fig:Scatter_demo}
\end{figure}

\begin{figure*}[!tbp]
  \centering
  \includegraphics[width=\textwidth]{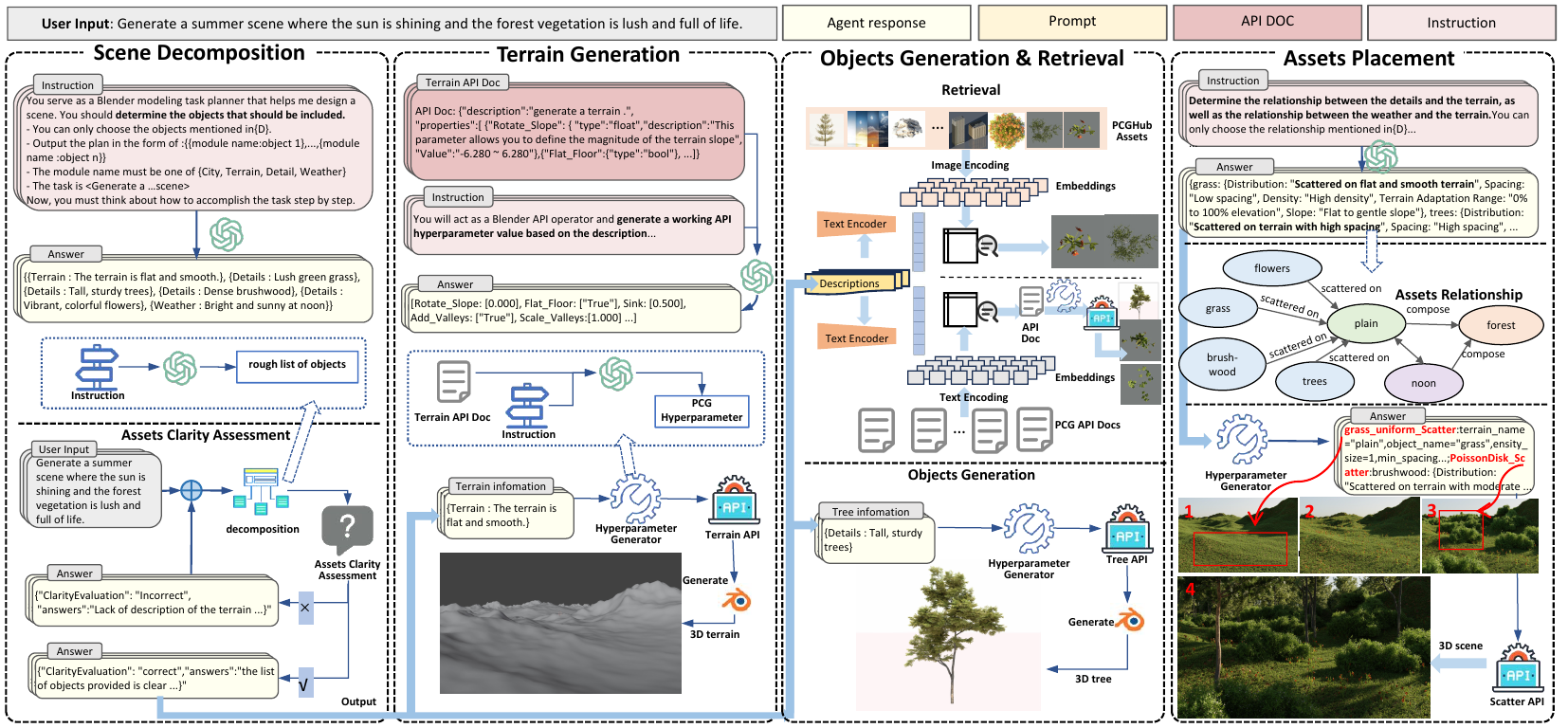}
  \caption{\mymethod{} framework converting user text input into diverse 3D scenes through four stages: scene decomposition stage, terrain generation stage, objects generation \& retrieval stage and assets placement stage.}
 \label{Framework}
 \vspace{-0.5cm}
\end{figure*}

\subsection{PCGHub}
For scene generation tasks, the diversity of scenes is intrinsically linked to the diversity of assets. Therefore, we introduce PCGHub, a platform that integrates a wide range of PCG modules and 3D assets. It provides detailed documentation and APIs for swift integration of varied PCG techniques, addressing the limitations of traditional methods and improving content realism. A summary of the extensive PCG modules available in PCGHub is provided in Table \ref{tab:pcghub_capabilities}. Further details are available in the supplement.

\noindent\textbf{Procedural Assets.} 
%To develop PCGHub, we assemble a comprehensive collection of 102 procedural assets and 11,284 high-quality 3D static assets, categorized into natural environments, architecture, biology, environmental impacts, and science fiction elements. These procedural assets are generated using Blender node graphs, which allow for the creation of a virtually infinite range of assets through adjustments to geometry and shader parameters. The assets are encapsulated as APIs and documented to facilitate invocation via LLMs. Specifically, 235 human-interpretable high-level APIs (e.g., the number of floors in a building) are derived from the 2,362 original node APIs, with functional descriptions, parameter details, API types, and names documented. Fig.\ref{fig:format} illustrates a detailed overview of the API documentation and the trees generated through the API.
To develop PCGHub, we assemble a comprehensive collection of 172 procedural assets, categorized into natural environments, architecture, biology, environmental impacts, and science fiction elements. These assets are generated using Blender node graphs, which allow the creation of a virtually infinite range of variations through adjustments to geometry and shader parameters. From the original 2,362 parameters, we extract 263 human-interpretable parameters and encapsulate them into APIs. Each API is comprehensively documented, including functional descriptions, parameter specifications, and types, to facilitate their utilization via language models.
 Fig. \ref{fig:tree} illustrates an example of API documentation, the corresponding APIs, and the resulting trees generated through these APIs. To further address the issue of limited PCG asset variety, we also collect 11,284 high-quality 3D static assets, enhancing the overall diversity and richness of the available assets.
 
 \noindent\textbf{Procedural Layout Generators.}
Complex patterns of arrangement and placement are inherent in various environments. To replicate these patterns, we predefine five types of layouts: scatter layout, which distributes assets randomly within a given area; grid layout, which arranges objects in a uniform grid pattern; linear layout, which arranges objects sequentially along a defined path, such as roads, rivers, or railways; nested layout, which organizes objects within a larger structure, such as buildings in a small neighborhood or attractions within a park; and area filling layout, which places objects according to specific rules to fill an entire designated area. For each layout type, we provide one or more corresponding procedural layout generators. By utilizing various objects for each layout type, we can generate a diverse range of scenes. Fig. \ref{fig:Scatter_demo} provides an example of the scatter layout, showcasing the effectiveness and diversity of procedural layout methods.

\subsection{PCGPlanner}
PCGPlanner utilizes the resources provided by PCGHub for efficient scene generation. As illustrated in Fig. \ref{Framework}, the fully automated scene generation process comprises the following stages:(1) \textit{Scene Decomposition}: Analyzes the scene requirements to identify the necessary assets; (2) \textit{Terrain Generation}: Constructs the foundational terrain and applies the appropriate materials; (3) \textit{Objects Generation \& Retrieval}: Involves generating or importing the assets required for the scene; (4) \textit{Asset Placement}: Utilizes diverse layout types and procedural generators to arrange assets within the scene.
\noindent\textbf{Systematic Template.} 
%To endow the LLMs with the capability for modeling scenes like a specialist, 
%We introduce a systematic template to prompt LLMs as agents to achieve the subtasks in a scene modeling task. For each agent, the prompt $P$ has a similar structure defined as $P_{\text{i}}$($R_{\text{i}}$, $T_{\text{i}}$, $D_{\text{i}}$, $F_{\text{i}}$, $E_{\text{i}}$), where i $\in$\{\textit{dispatch, specialist, retrieval, execution}\} distinguishes the responsibility of agents. The constituents of the prompt are  defined as follows:
To endow the LLMs with the capability for modeling scenes, We introduce a systematic template to prompt LLMs in a scene modeling task. For each agent, the prompt $P$ has a similar structure defined as $P_{\text{i}}$ ($R_{\text{i}}$, $T_{\text{i}}$, $D_{\text{i}}$, $F_{\text{i}}$, $E_{\text{i}}$), where i corresponds to different subtasks . The constituents of the prompt are  defined as follows:
\begin{itemize}
    \item[] \hspace{-0.25em} {\small $\bullet$} \textbf{Role} Each agent is given a specific role R which describes its responsibility in the scene generation process. 
    \item[] \hspace{-0.3em} {\small $\bullet$} \textbf{Task} T gives a detailed explanation of the goals for the agent. In the meanwhile, the constraints are also expounded executing these tasks.
    \item[] \hspace{-0.3em} {\small $\bullet$} \textbf{Document} At each step, the agent is prompted by a knowledge document according to their task. We denote D as the collection that contains all the knowledge documentation pre-defined in PCGHub. 
    \item[] \hspace{-0.3em} {\small $\bullet$} \textbf{Format} F denotes the output format for each agent. To precisely and concisely convey information between agents, the output format of each agent is strictly defined. 
    \item[] \hspace{-0.15em} {\small $\bullet$} \textbf{\hspace{0.1em}Examples} To help agents respond strictly following the format, we demonstrate several examples E for reference in each prompt.
\end{itemize}
The systematic template guides LLMs to produce accurate outputs, enhancing the success rate and executability rate.

\noindent\textbf{Scene Decomposition Stage.}
As illustrated in Fig. \ref{Framework}, this stage converts user textual descriptions into a rough list of objects for the target scene. The detailed object requirements and their corresponding procedural modules are outlined in the knowledge documentation $D_{decomposition}$. The process can be formulated as follows:
\[
\{o_1, \dots, o_n\} \leftarrow LLM_{decomposition}(q, P_{decomposition})
\]
where $q$ is the user input, and $o_i$ represents an object with its associated modules and descriptions structured as \{\{\textit{module: description}\}, ... \{\textit{module: description}\}\}.

Additionally, recognizing that user may omit essential details, we incorporate an Assets Clarity Assessment to identify and resolve ambiguities by actively querying users to obtain missing details for a complete preliminary plan.

\noindent\textbf{Hyperparameter Generator.}
The Hyperparameter Generator (HyperparamGen) performs as an executor, which converts textual descriptions into PCG parameters and applies these parameters to APIs. Within PCGHub, each API is defined as a Python function with multiple predefined parameters. For each API, a corresponding knowledge document $D_{\alpha}$ $\in$ $D$ is provided to the Hyperparameter Generator. This document contains essential details required to generate the appropriate hyperparameters for executing the API. The hyperparameter generation process is formulated as follows:
%\begin{equation}
%\alpha^* \leftarrow \text{HyperparamGen}(\alpha, P_{\text{HyperparamGen}}, o_j)
%\end{equation}
\begin{equation}
\alpha^* \leftarrow \text{HyperparamGen}(\alpha, P_\alpha, o_i)
\end{equation}
where \(\alpha\) represents the initial hyperparameters and $\alpha^*$ denotes the optimized parameters. In cases where the description may include incomplete parameter information, the Hyperparameter Generator is instructed to automatically select feasible parameters from available options based on the other parameters. The inferred parameters are utilized to execute the corresponding APIs in the Blender Python environment, completing all tasks efficiently.

\noindent\textbf{Terrain Generation Stage.}
This stage is responsible for terrain generation within the scene. Unlike other assets, terrain lacks distinct characteristics that can be effectively identified using CLIP \cite{CLIP}. Thus, a specialized terrain generation PCG is employed to ensure accurate and detailed control. As shown in the second panel of Fig. \ref{Framework}, the PCG provides extensive control over terrain attributes, including geometry and material properties. It enables precise adjustments to terrain slope, elevation, and features such as valleys, ensuring that the generated terrain meets specific scene requirements with both flexibility and accuracy.

\noindent\textbf{Objects Generation \& Retrieval Stage.}
Based on the detailed object descriptions provided by the Scene Decomposition Stage, we need to search within PCGHub for assets to directly import or suitable APIs to generate procedural assets. To achieve accurate retrieval, the Objects Generation \& Retrieval Stage manages this process by encoding these descriptions into embeddings for retrieval. 
A pre-trained CLIP model is employed: text-to-text retrieval is used for procedural assets APIs, and text-to-image retrieval is used for 3D static assets. As depicted in the third panel of Fig. \ref{Framework}, each static asset in PCGHub is represented by a 768-dimensional vector derived from its rendering image. These vectors are compared with the input description embedding, and one of the top five most similar results is randomly selected based on cosine similarity and imported into the Blender scene. 
%\begin{equation}
%(\alpha, \gamma) \leftarrow \text{Retrieval}(o_j)
%\end{equation}
%where $\alpha \in \mathcal{A} $ represents the retrieved API from the API collection $\mathcal{A}$, and $\gamma \in \Gamma$ is the retrieved static asset from asset collection $\Gamma$.
\begin{equation}
\text{Asset}_j \leftarrow \text{Retrieval}(o_j)
\end{equation}
where $\text{Asset}_j \in \Gamma$ represents the retrieved static asset from the asset collection $\Gamma$. For text-to-text retrieval, the most relevant API is also selected by its functional description embedding. If the retrieval result is an asset, it will be directly imported into the scene project. On the other hand, the retrieved API is to be passed to the Hyperparameter Generator.

\noindent\textbf{Assets Placement.}
In large-scale scene generation tasks, asset placement is inherently complex. It involves not only identifying spatial relationships between objects but also determining how these relationships should be represented. To address this, we propose a method based on asset relationships to guide asset placement. As illustrated in the fourth panel of Fig. \ref{Framework}, the LLM, guided by the predefined layout types, selects the appropriate layout for each object. Subsequently, the LLM defines the spatial relationships among assets in the scene. Each asset relationship pair is processed by the Hyperparameter Generator, which generates the necessary parameters for asset placement. These parameters are then utilized by the Procedural Layout Generators to arrange the assets accordingly in Blender.

\section{Experiments}
The goals of our experiments are threefold: (i) to verify the capability of \mymethod{} for generating photorealistic large-scale scenes, including nature scenes and cities, (ii) to demonstrate the effectiveness of \mymethod{} for personalized editing, such as adding or changing,(iii) to compare different LLMs on the proposed benchmark.

\subsection{Benchmark Protocol}
\textbf{Dataset.} To evaluate the effectiveness of proposed \mymethod{}, we use GPT-4 to generate high-quality 50 scene descriptions, 50 asset descriptions, and 20 asset editing descriptions. The scene descriptions involve natural scenes and cities. Then, we feed them to our \mymethod{} to generate corresponding models, which are used to perform quantitative and qualitative comparisons.

\noindent\textbf{Models.} When generating and editing the 3D scenes, we adopt the leading GPT-4 as the large language model with its public API keys. To ensure the stability of LLM's output, we set the decoding temperature as 0.

\noindent\textbf{Metrics.} We use Executability Rate (ER@1) and Success Rate (SR@1) to evaluate LLMs on our \mymethod{}. The former measures the proportion of proposed actions that can be executed, and the latter is used to evaluate action correctness \cite{chen2021evaluating}.
Moreover, to quantify aesthetic quality, we adopt a unified judgment standard as a reference. We divide the aesthetics of generated scenes into five standards: Poor (1-2 points)/Below Average (3-4 points)/Average (5-6 points)/Good (7-8 points)/Excellent (9-10 points). We enlisted 35 volunteers to assess the quality of our generation, including 5 PCG experts. We compute average score (AS) and average expert score (AES) to evaluate the effectiveness of our method.

\subsection{Main Results}

\begin{figure}[!tbp]
\centering
\includegraphics[width=1\linewidth]{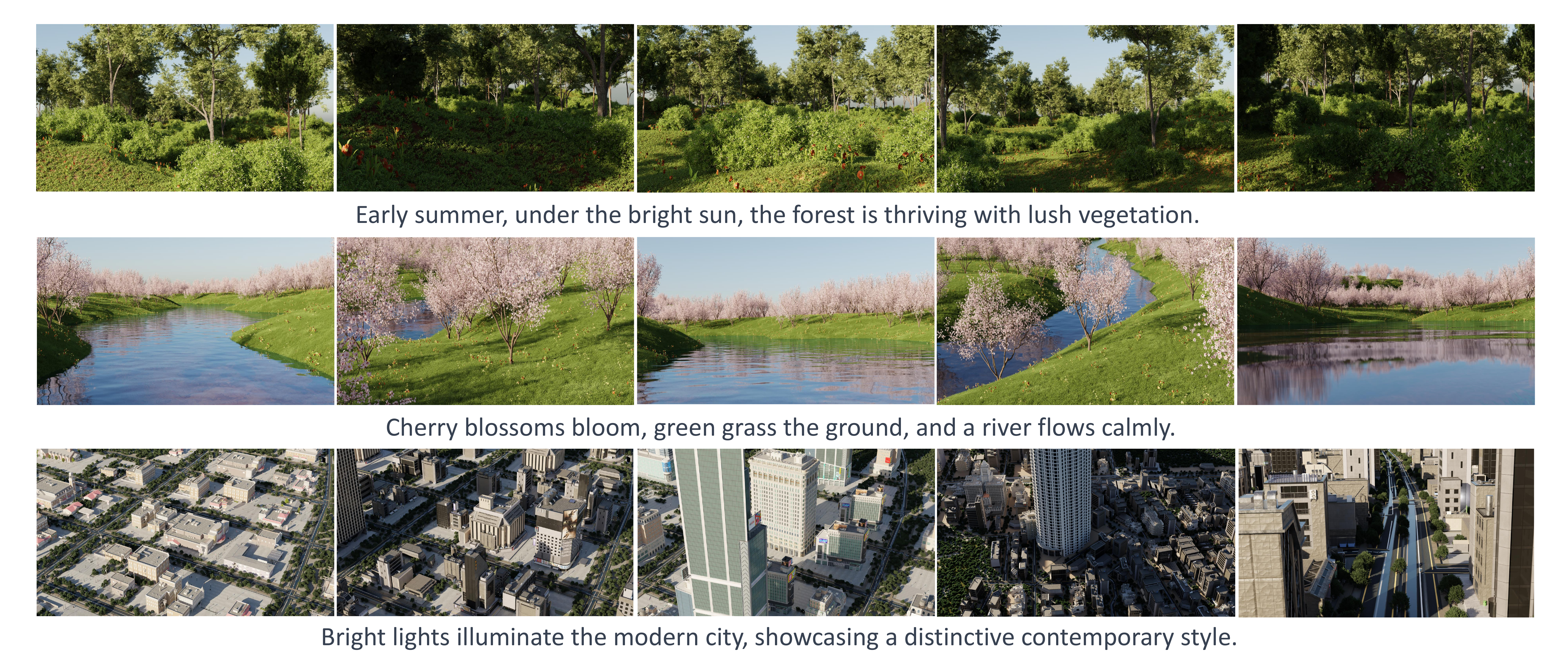}
%\vspace{-0.3cm}
\caption{Visualization of the generation quality for large-scale scenes and cities.}
\vspace{-0.5cm}
\label{fig:example}
\end{figure}

\begin{figure}[!tbp]
\centering
\includegraphics[width=1\linewidth]{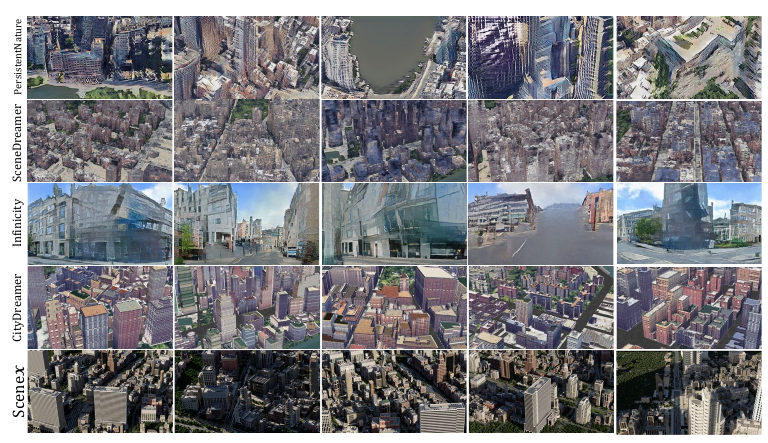}
% \vspace{-0.5cm}
\vspace{-20pt} 
\caption{Comparative results on city generation. }%Issues with unreasonable geometry are observed in previous works, while our method performs well in generating realistic large-scale city scenes.}
\vspace{-0.6cm}
\label{fig:city_example}
\end{figure}

\noindent\textbf{Generation Quality.} We begin by showcasing several examples of our \mymethod{} for generating large-scale nature scenes and unbounded cities. The results are shown in Fig. \ref{fig:example}. From the results, we can observe that the proposed \mymethod{} can produce highly realistic scenes in both natural scenes and cities. Moreover, the generated content is correctly corresponding to the provided texture descriptions. These demonstrate the power and effectiveness of our proposed LLM-driven automatic 3D scene generation framework.
Fig. \ref{fig:city_example} shows the qualitative comparison results between the learning-based methods for city scene reconstruction work and \mymethod{}.  From the results, we can observe that learning-based methods commonly suffer similar problems: low 3D consistency and building structural distortions. For example, PersistanNature \cite{chai2023persistent} and InfiniCity \cite{InfiniCity} both appear to have severe deformation in the whole scene level. SceneDreamer \cite{SceneDreamer} and CityDreamer \cite{xie2023citydreamer} have better structure consistency, but the building quality is still relatively low. These factors limit their large-scale application in industry. In comparison, the proposed \mymethod{} generates highly realistic and well-structured urban scenes without the issues of structural distortions and layout defects compared to learning-based methods. These results demonstrate the effectiveness of \mymethod{} for large-scale city scene generation.

\begin{table}[!tbp]
\centering
\small
\setlength{\tabcolsep}{9pt}
%\resizebox{\columnwidth}{!}{ % 使用 \columnwidth 替代 \textwidth
\begin{tabular}{lcc}
\toprule[0.7pt]
Method & AS  & AES   \\ 
\midrule[0.7pt]
Magic 3D    \cite{Magic3D}  & 4.48 & 3.50 \\
DreamFusion \cite{DreamFusion}  & 4.55 & 3.60 \\
Text2Room \cite{Text2Room}  & 5.73 & 6.10 \\
CityDreamer \cite{xie2023citydreamer} & 5.47 & 6.80 \\
Infinigen \cite{Infinigen} & 5.42 & 6.00 \\
3D-GPT \cite{3D-GPT}  & 4.94 & 6.20 \\
WonderJ \cite{yu2023wonderjourney}  & 5.28 & 6.00 \\
\textbf{\mymethod{} (Ours)} & \textbf{7.83} & \textbf{7.70} \\ \bottomrule[0.7pt]
\end{tabular}
%}
\vspace{-7pt} 
\caption{Comparative analysis of average score (AS) and average expert score (AES). %The best results are highlighted in \textbf{bold}. %Highlighting the Superiority of Our Approach in the Domain of Aesthetic Evaluation
}
\label{tab:Aesthetic}
\vspace{-0.4cm}
\end{table}

\noindent\textbf{Aesthetic Evaluation.} To better evaluate the generation quality of \mymethod{}, we collect the results of related works involving text-to-3D work and Blender-driven 3D generation. These results are subjected to aesthetic evaluation by a panel comprising of 35 voluntary contributors and 5 experts in 3D modeling, with the scoring criteria outlined in Section 4.1. As shown in Table \ref{tab:Aesthetic}, our scores for AS and AES surpass the second-highest scores by 2.10 and 0.9 points, respectively. Compared to the other works, our project reaches a good level, indicating the high generation quality of \mymethod{}.

To evaluate the consistency between text inputs and the generated assets, we calculate the CLIP similarity between input text and rendered images. To better illustrate the results, we utilize three different CLIP models for testing, including ViT-L/14 (V-L/14), ViT-B/16 (V-B/16) and ViT-B/32 (V-B/32), respectively. The detailed results are displayed in Table \ref{tab:clip-sim}. We compare representative text-to-3D approaches (\textit{e.g.} WonderJ \cite{yu2023wonderjourney}, Text2Room \cite{Text2Room}, and DreamFusion \cite{DreamFusion}) and Blender-driven 3D generation works (\textit{e.g.} BlenderGPT, 3D-GPT \cite{3D-GPT}, and SceneCraft \cite{hu2024scenecraft}). Although the similarity scores of text-to-3D methods are higher, it is reasonable because their training or optimization includes the text-to-image alignment process. Compared to the Blender-driven 3D generation works, \mymethod{} achieves the highest score, indicating its capability to accurately execute the input prompts and generate results.

\begin{table}[!tbp]
\centering
\small

\setlength{\tabcolsep}{2pt}
%\resizebox{\columnwidth}{!}{
\begin{tabular}{lccc}
\toprule[0.7pt]
%\multicolumn{2}{l}{}              & \multicolumn{3}{c}{CLIP SIM}         \\ %\cmidrule{2-4}
              & V-L/14 & V-B/16 & V-B/32 \\ 
              \midrule[0.7pt]
{WonderJ~\cite{yu2023wonderjourney}} & 18.78  & 25.70    & 25.45    \\
{Text2Room~\cite{Text2Room}}     & 23.51  & 30.10   & 29.29    \\
{Magic 3D~\cite{Magic3D}}      & 27.86   & 31.78   & 31.94    \\
{DreamFusion~\cite{DreamFusion}} & \textbf{29.40} & \textbf{35.37} & \textbf{31.60} \\ \midrule
{BlenderGPT}    & 21.23    & 25.65    & 26.19    \\
{3D-GPT~\cite{3D-GPT}}        & 18.67    & 25.80  & 25.59    \\
{SceneCraft~\cite{hu2024scenecraft}}    & 22.04  & 25.82    & 25.30    \\
{\textbf{\mymethod{} (Ours)}}        & \textcolor{gray}{22.82}    & \textcolor{gray}{27.82}    & \textcolor{gray}{26.89}    \\ \bottomrule[0.7pt]
\end{tabular}
%}
\vspace{-7pt} 
\caption{Assessing prompt-rendered result similarity with various models. %Higher CLIP Similarity Observed in CLIP-optimized Methods Compared to LLM-driven Ones
}
\label{tab:clip-sim}
\vspace{-5pt} 
\end{table}

\begin{table}[!tbp]
\centering
\footnotesize
\setlength{\tabcolsep}{6pt}
\resizebox{\columnwidth}{!}{
\begin{tabular}{l@{\hspace{5pt}}c@{\hspace{5pt}}c@{\hspace{5pt}}c@{\hspace{5pt}}c@{\hspace{5pt}}c@{\hspace{5pt}}c@{\hspace{5pt}}c@{\hspace{5pt}}c}
\hline
\multirow{2}{*}{} & \multicolumn{2}{c}{150m $\times$ 150m} &  & \multicolumn{2}{c}{500m $\times$ 500m} &  & \multicolumn{2}{c}{2.5km $\times$ 2.5km} \\ \cline{2-3} \cline{5-6} \cline{8-9} 
          & Scene & City  &  & Scene & City &  & Scene & City             \\ \hline
Human     & 1h    & 40min &  & 3h    & 4h   &  & -     & \textgreater{}3w \\
Infinigen  & 14min & -     &  & -     & -    &  & -     & -                \\
\textbf{\mymethod{}}    & \textbf{2min} & \textbf{1.5min} &  & \textbf{10min} & \textbf{6min} &  & -         & \textbf{20h}         \\ \hline
\end{tabular}
}
\vspace{-7pt} 
\caption{Comparing the time required for natural scene generation and city generation at different terrain scales.}
\label{tab:time}
\vspace{-0.6cm}
% \vspace{-60pt} 
\end{table}

\noindent\textbf{Efficiency Evaluation.} To illustrate the efficiency of \mymethod{}, we provide the time required of our method compared with Infinigen \cite{Infinigen} and human craft.
The experiments are performed on a server equipped with dual Intel Xeon Processors (Skylake architecture), each with 20 cores, totaling 80 CPU cores. Additionally, we consult 3D PCG experts to determine the time needed to construct the same natural and urban scenes. The comparison results are shown in Table \ref{tab:time}. From the results, we can observe that \mymethod{} is 7 times faster than Infinigen \cite{Infinigen} in generating a scene with 150m $\times$ 150m size. We also provide nearly 30 times faster than human experts creating a large-scale city by hand. This demonstrates the impressive efficiency of \mymethod{} in both large-scale natural scene and urban scene generation.

\noindent\textbf{Personalized Editing Results.} To demonstrate the capability of our method for personalized editing, we conduct experiments on 3D asset generation guided by users' instructions. 
The results are shown in Fig. \ref{fig:tree_example}. It is evident that the changes in the edited text are closely related to the modifications in 3D assets. \mymethod{} demonstrates a versatile, highly controllable, and personalized editing ability by manipulating 3D assets from various perspectives. These results demonstrate that our method supports user-instruction editing, significantly reducing the difficulty of 3D asset generation and accelerating the industrial production process.

\begin{table}[!tbp]

\setlength{\tabcolsep}{4.7pt}
\resizebox{\columnwidth}{!}{
\begin{tabular}{c|cccccccc|cc}
\toprule[0.7pt]
 & Task & Document &  Examples & Role  & ER@1  & SR@1 \\ 
\midrule[0.7pt]
1   & \checkmark  &   &   &   & 16.00 & 25.00 \\
2   &  & \checkmark   &  &  & 42.00 & 47.62\\
3   & \checkmark  & \checkmark & & & 84.00 &  71.43\\
4   &  & & \checkmark & & 92.00 & 73.91\\
5   & \checkmark & & \checkmark & & 92.00 &  76.09\\
6   &  & \checkmark   & \checkmark     &  & 92.00 & 76.09\\
7   & \checkmark  & \checkmark  & \checkmark  & & 92.00 & 78.26 \\
8   &   & \checkmark   &  \checkmark & \checkmark  & 94.00 & 78.72\\
9   & \checkmark  & \checkmark  & \checkmark  & \checkmark    &  \textbf{94.00} & \textbf{80.85}\\ 
\bottomrule[0.7pt]
\end{tabular}
}
\vspace{-7pt} 
\caption{Results of different prompt components for tree asset generation. Examples, Role, Document and Task represent four integral components of the prompt template.}
\label{tab:Ablation studies}
\vspace{0.1cm}
% \vspace{-50pt} 
\end{table}

\begin{figure}[!tbp]
\centering
\includegraphics[width=\linewidth]{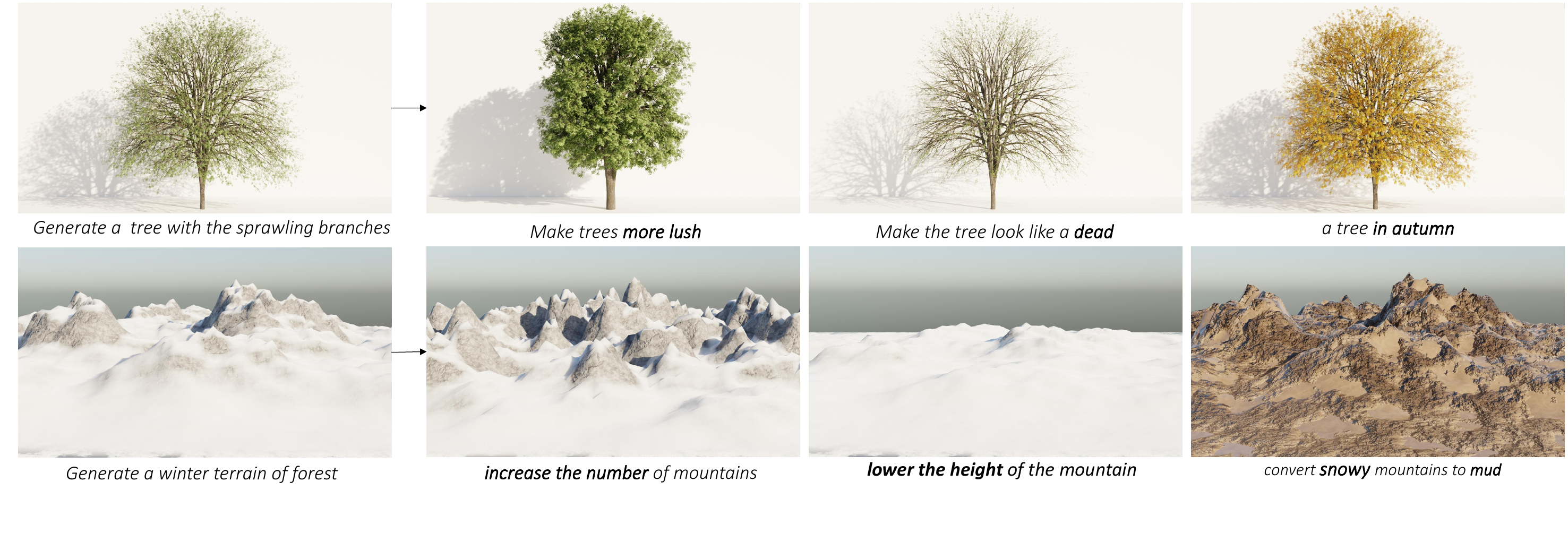}
\vspace{-0.7cm}
\caption{Visualization of the personalized editing results on different scenes.}
\label{fig:tree_example}
\vspace{-0.6cm}
\end{figure}

\subsection{Ablation Study}

To analyze the impact of various components within the systematic template, we conduct an ablation study based on the tree plugin in PCGHub. During the experiment, we utilize the dataset from Section 4.1 for testing, maintaining a consistent format. We incrementally add or remove different parts of the systematic template, using ER@1 and SR@1 as metrics to observe the impact of various components on the system. The results are shown in Table \ref{tab:Ablation studies}. It is evident from the results that augmenting the Task, Document, Examples and Role components contributes to an increase in ER@1 and SR@1. Among these, the inclusion of the Example component results in the most significant improvement, resulting in a maximum increase of 76.00\% and 51.09\% in ER@1 and SR@1 respectively. Conversely, the Role component has the least impact, with maximum increases of 2.00\% and 2.59\% in ER@1 and SR@1 after its addition. These results suggest that the setting of Task significantly impacts the performance of LLMs. By clarifying the goals and requirements of the task, the system can better understand the user's intent and better meet the needs when generating output. Examples play an important role in designing proxy prompts. By providing concrete examples, the system can better understand the user's intent and needs and produce high-quality output related to the input text. %The inclusion of documentation is also important for designing proxy prompts. 
Documentation provides background information that can help the system better meet user expectations when generating output.

\subsection{Comparing with Different LLMs}
To investigate the performance of different variants of large language models (LLM) in \mymethod{}, we test public LLM APIs like gpt-3.5-turbo, gpt-4 and several external open-source LLMs in this subsection. To ensure the stability of LLM outputs, we set the temperature of LLM to 0 for all experiments.
We conduct experiments on 3D scene and asset generation based on 50 scenario descriptions and 50 object descriptions in Sec. 4.1. 
The results are presented in Table \ref{tab:ERSR}.
It is evident that gpt-4 delivers the superior performance, with Mistral closely following as the second-best. Due to its performance and lower hardware requirements, the open-source Mistral is a highly appealing option. When compared to asset generation, the executability and success rates noticeably decline during the generation of large-scale natural scenes, a trend that can be attributed to the increased task complexity. In particular, as the number of components involved in the system expands, the LLM may face challenges in maintaining their accuracy. Nonetheless, our method exhibits consistent performance across different LLMs, maintaining high levels of executability and success rates.

\begin{table}[!tbp]

\centering
\setlength{\tabcolsep}{4pt} % 调整列间距以适应内容
\resizebox{\columnwidth}{!}{
\begin{tabular}{lcccccl}
\toprule
\multirow{2}{*}{Model} & \multicolumn{2}{c}{Scene Generation} & & \multicolumn{2}{c}{Asset Generation} & \\ 
\cline{2-3} \cline{5-6}
 & ER@1 & SR@1 & & ER@1 & SR@1 & \\ 
\midrule
Llama2-7B\cite{touvron2023llama} & 30.00 & 53.33 & & 38.00 & 57.89 & \\
Llama2-13B\cite{touvron2023llama} & 44.00 & 59.09 & & 54.00 & 66.66 & \\
Mistral\cite{jiang2023mistral} & 76.00 & 68.42 & & 94.00 & 85.11 & \\
Gemma-2B\cite{gemmateam} & 6.00 & 33.33 & & 22.00 & 45.45 & \\
Gemma-7B\cite{gemmateam} & 36.00 & 55.56 & & 68.00 & 73.53 & \\
GPT-3.5-turbo\cite{brown2020language} & 66.00 & 60.60 & & 82.00 & 82.93 & \\
GPT-4\cite{openai2023gpt4} & \textbf{86.00} & \textbf{86.05} & & \textbf{96.00} & \textbf{85.42} & \\
\bottomrule
\end{tabular}
}
\vspace{-7pt} 
\caption{Comparing the performance of different language models in natural scene generation and city generation.}
\label{tab:ERSR}
% \vspace{-20pt}
\vspace{-0.6cm}
\end{table}

\section{Conclusion}
% 我们正在不断吸纳更多的PCG，我们尽可能最完整最全面的生成的框架、平台、社区，把100多个插件体现出来
In this paper, we introduce \mymethod{}, an advanced framework for large-scale scene generation that automatically creates high-quality procedural models from textual descriptions. The framework consists of two key components: PCGHub and PCGPlanner. PCGHub includes an extensive repository of accessible procedural assets and thousands of meticulously crafted API documents. Meanwhile, the PCGPlanner can generate executable actions for Blender to create controllable and precise 3D assets following user instructions. \mymethod{} can generate a 2.5 km × 2.5 km city with accurate layout and geometric structures, dramatically reducing production time from two weeks of professional work to just a few hours for an ordinary user. Our method demonstrates significant capabilities in controllable scene generation and editing, validated through extensive experiments. 
%In addition to refining existing domains, we are actively expanding into new key areas, such as the generation of dynamic elements and the construction of interactive 3D scenes. We keep adding new PCG modules to support these advancements, ensuring that \mymethod{} remains at the forefront of procedural generation capabilities.

\noindent\textbf{Limitations.} %Although the proposed \mymethod{} is effective for the LLM-driven procedural large-scale scene generation,
%it still faces several limitations: 1) The performance of our method relies on the pretrained LLM model. This reliance may constrain the framework from generalizing to a wider range of applications. 2) We have collected a limited number of assets and APIs, which limits the variety of generated scenes and degree of action space. In the future, it is promising to fine-tune the LLMs with professional modeling knowledge and empower them to be a professional modeling assistant. Collecting more resources for PCGHub can also enhance the scene fidelity and aesthetics.
Although \mymethod{} effectively enables LLM-driven procedural large-scale scene generation, it has certain limitations: 1) Dependence on the pretrained LLM model may restrict its generalization across applications. 2) It is constrained by the inherent algorithms and rules of existing PCG modules. Expanding PCGHub's resources could further enhance scene diversity and quality.

\section{Acknowledgments}
This work was supported in part by the National Natural Science Foundation of China (No. U21B2042, No. 62320106010, No. 62072457), and the 2035 Innovation Program of CAS, and the InnoHK program.

\bibliography{aaai25}

\clearpage
\appendix
\section{Supplementary Materials}

In this paper, we propose an advanced \mymethod{} for large-scale scene generation according to the user's instructions. The reproducible code is attached to the supplementary material, and some other important details are as follows.

\subsection{A. PCGHub Management}
PCGHub offers an extensive collection of PCG modules and 3D asset resources designed to meet the demands of generating large-scale scenes. The management of PCGHub involves several key processes to ensure efficient operation and seamless integration of resources.
% \subsubsection{A.1. PCGHub Resources}
%PCGHub offers an extensive collection of PCG modules and 3D asset resources designed to meet the demands of generating large-scale scenes. 
%Our platform is structured around four key capabilities: Materials and Shading, Modeling, Rendering, and Asset Placement. For generation-based PCG, We provide a diverse selection of PCG modules, including terrain, trees, flowers, rocks, and buildings. In the realm of material-based PCG, we provide editable materials such as textures for plant leaves, roots, terrain surfaces, sky, metals, and more. 
%Our asset placement-based PCG resources encompass various methods, including slope-based, height-based, object-based, and geometric space-based placements, ensuring precise and contextual placement of assets. Rendering-based PCG resources are designed to optimize rendering engines, settings, and overall visual output. Additionally, PCGHub includes a comprehensive library of 3D assets, spanning categories like buildings, vegetation, urban infrastructure, and textures, sourced from a wide array of online resources, providing users with a robust toolkit for scene generation.

\subsubsection{A.1.Encapsulation Standard}
To optimize the usage of PCG plugins, we need to simplify each plugin. The simplification rules are as follows:
\begin{itemize}
\item Generation Plugins: Extract APIs related to grid generation and remove irrelevant APIs, ensuring the ability to independently generate the corresponding 3D grid.
\item Material Plugins: Extract APIs providing material variations and remove irrelevant APIs, ensuring the ability to independently modify the corresponding material.
\item Placement Plugins: Extract APIs related to placement functionality and remove irrelevant APIs, ensuring the ability to place objects in the current scene.
\item Rendering Plugins: Extract APIs providing scene rendering capabilities and remove irrelevant APIs, ensuring the ability to perform rendering tasks for the entire 3D scene.
\end{itemize}

\subsubsection{A.2.PCG Encapsulation} 
% PCG封装用于标准化和打包每个程序模块,从而确保在不同场景和应用中的一致使用。这种封装便于程序模块的替换和组合,从而增强平台内的模块化和可重用性。
PCG Encapsulation is a crucial process in PCGHub's development, designed to integrate diverse PCG plugins—especially Blender node graph-based plugins—by establishing a standardized framework for their systematic utilization.

\begin{figure}[!tbp]
\centering
\includegraphics[width=1\linewidth]{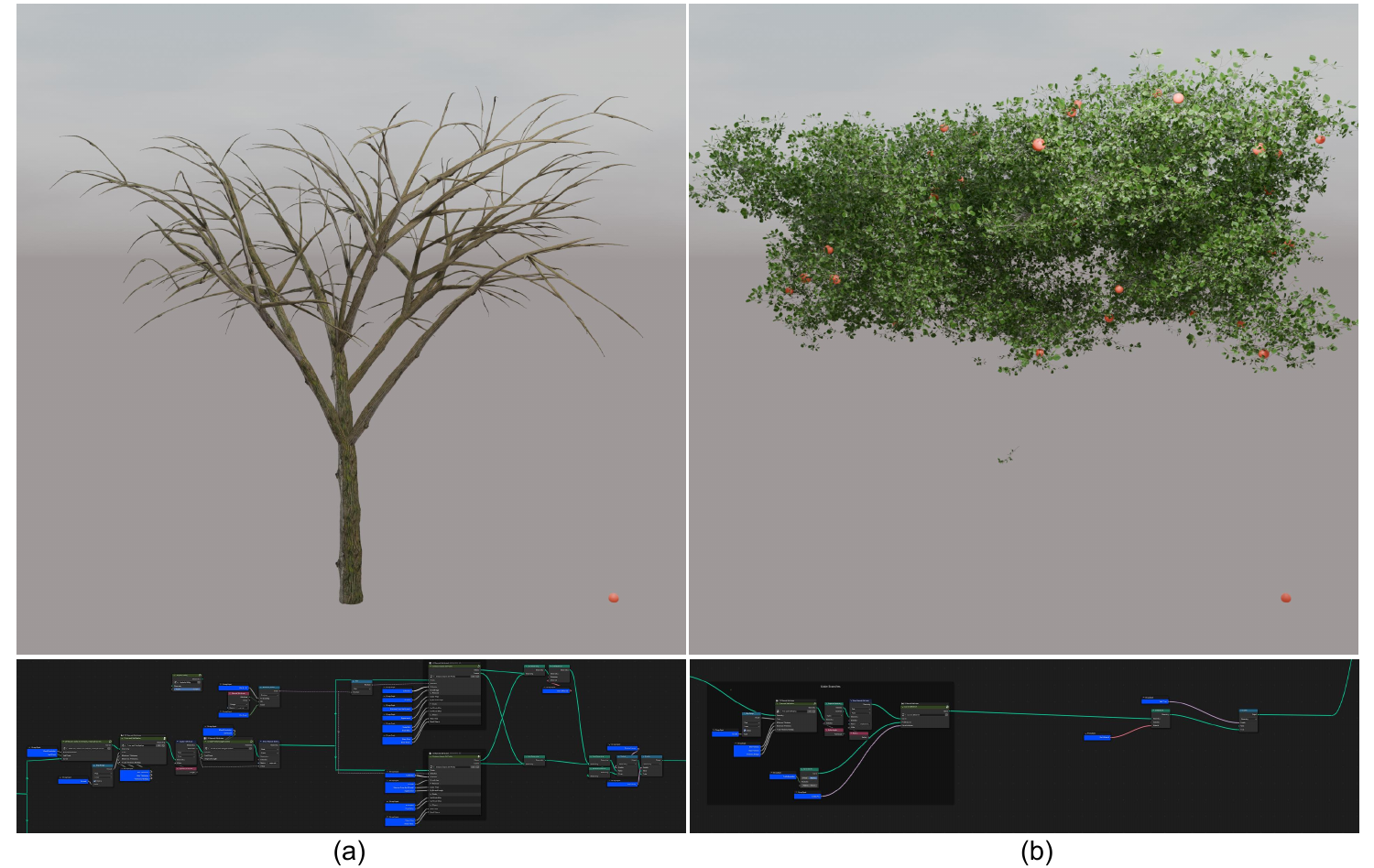}
%\vspace{-0.3cm}
\caption{Node Graph Visualization: From Core Logic to Functional Outputs in PCG Modules. (a) Generate the root of the tree and its corresponding node graph. (b) Generate leaves and its corresponding node graph.}
\vspace{-0.2cm}
\label{fig:tree_node_graph}
\end{figure}

The encapsulation process meticulously examines each PCG plugin to extract its core program logic and input-output parameters. This approach ensures that each plugin's essential functionality is preserved while enabling interoperability across different procedural generation tasks. The resulting encapsulated modules are stored in the blend file format, providing a standardized, reusable resource. Fig.\ref{fig:tree_node_graph} illustrates the core program logic through a node graph visualization, showing the generation of tree roots and leaves, each with its corresponding node graph.

\subsubsection{A.3.API Collection}
% These PCG plugins are predominantly categorized into two types: one based on geometric node groups and the other based on Python code.It is necessary to gather API-related information from these PCG plugins, including API names, API function descriptions, API parameter descriptions, API parameter value ranges, and API parameter default values. For plugins based on geometric node groups, in order to obtain their original APIs, we need to test the geometric node groups of the plugins in Blender software and document their original APIs along with the corresponding information. For plugins based on Python code, we need to test the source code and record the functions within the plugin code as the original APIs, while also documenting the relevant information for these functions.
% After the core logic and I/O parameters are established, the plugins are encapsulated as PCG modules and stored in blend file format. Following this, Python code is used to document each module's name, description, inputs, corresponding input descriptions and types, outputs, and constraints. This comprehensive documentation ensures that each PCG module is readily deployable across diverse procedural content generation tasks. As shown in fig. \ref{fig:code},...
PCG plugins are divided into two categories: those using geometric node groups and those based on Python code. To document these plugins, the following approach is taken:
For plugins based on geometric node groups, we test these node groups in Blender to gather API details, including names, function descriptions, parameter descriptions, value ranges, and default values.
For Python-based plugins, we analyze the source code to document functions, including names, purposes, parameters, ranges, default values, and constraints.

After defining the core logic and I/O parameters, the plugins are encapsulated as PCG modules in blend file format. Each module is then documented using Python code, covering the module’s name, description, inputs, input descriptions and types, outputs, and constraints. This documentation ensures each PCG module is effectively deployed for procedural content generation tasks. As shown in Fig. \ref{fig:code}, the standardized Python template for documenting PCG modules streamlines and details each module, facilitating its integration into various generation workflows.

\begin{figure}[!tbp]
\centering
\includegraphics[width=1\linewidth]{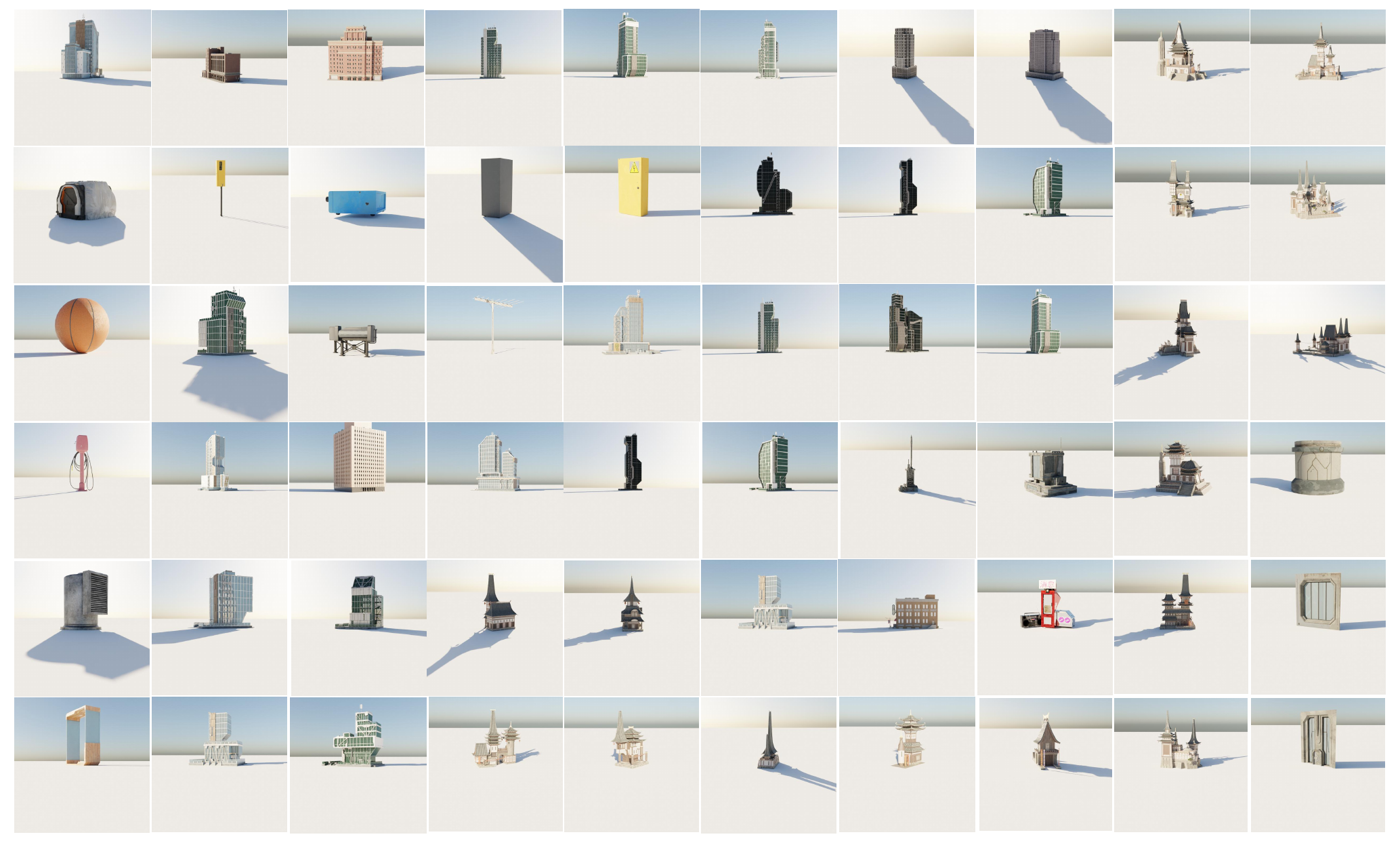}
%\vspace{-0.3cm}
\caption{Rendered building asset images for textual matching in asset retrieval.}
\vspace{-0.5cm}
\label{fig:Rendered_building_asset}
\end{figure}

\begin{figure*}[!tbp]
\centering
\includegraphics[width=\textwidth]{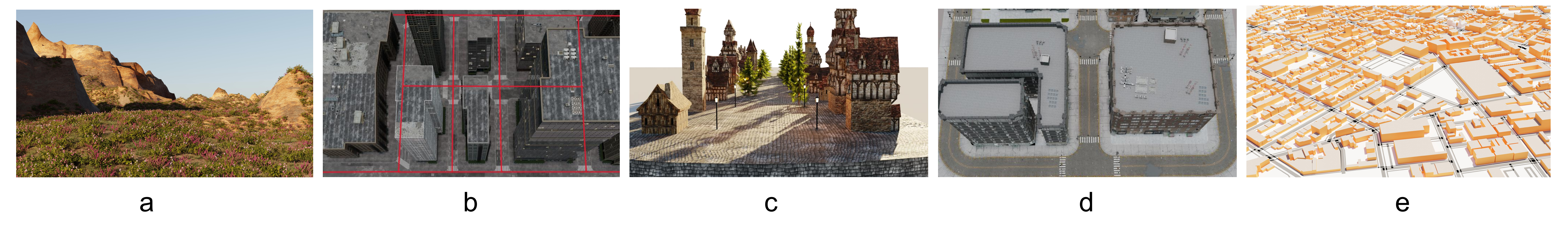}
\vspace{-0.73cm}
\caption{Examples of procedural layout generators—point distribution, grid-based, linear, nested, and region-filling.}
\label{fig:layout_demo}
\vspace{-0.3cm}
\end{figure*}

%\subsubsection{A.2.PCGHub Protocol}
%\subsubsection{A.3.Version Control}
\subsubsection{A.3.Rendering for Retrieval}
The Rendering for Retrieval process automates the visualization of assets to facilitate their effective retrieval and integration into procedural content generation workflows. The procedure begins by computing the bounding box of a given 3D object, which is used to determine the object’s geometric center and form factor. These geometric parameters are essential for accurately positioning a tracking target within the scene. Specifically, the code generates an empty object at the asset's geometric center, which serves as the focal point for subsequent camera configurations.
Once configured, the camera is set as the active camera for rendering, and the system generates images that faithfully represent the asset's key features. These images are then used for asset retrieval, enabling accurate matching with textual descriptions. As illustrated in Fig. \ref{fig:Rendered_building_asset}, this process applies to building assets, generating rendered images that support accurate textual matching and retrieval within the system.
% \subsection{B. PCGHub Resource}
% \subsubsection{B.1. Procedual Assets}
% \subsubsection{B.2. Static Assets}
%\subsubsection{B.3. Asset Stylization}

% \noindent\textbf{}

\subsection{B. More Details}
\subsubsection{B.1. Algorithm}
In this section, we present the pseudocode for the \mymethod{} framework implementation workflow, revealing the logical relationships across its stages. The process begins with a loop that checks the termination condition, halting execution when \texttt{End\_flag} is True. Within each iteration, the workflow initiates by generating a set of rough plan objects $\{o_1, \ldots, o_n\}$ from the Scene Decomposition stage, using input \texttt{q} and Prompt. Subsequently, each object is retrieved into the scene, followed by a RELATIONSHIP analysis to identify interactions among objects. Finally, asset placement is performed based on these relationships, continuing until all placement tasks are completed and \texttt{End\_flag} is set to True.

\subsubsection{B.2. Procedural Layout}
PCGHub's Procedural Layout Generators create diverse spatial configurations through five predefined layout types: Point Distribution, Grid-Based, Linear, Nested, and Region-Filling. These layouts model different real-world patterns, allowing for the simulation of complex environments.
As illustrated in Fig. \ref{fig:layout_demo}, the image demonstrate five examples of these layout types.

\subsubsection{B.3. Examples of Prompt Details}
To provide a more intuitive demonstration of how prompts are designed and executed, we are presenting a series of examples. In Fig. \ref{fig:terrain_example}, \ref{fig:placement_example}, \ref{fig:tree_example}, and \ref{fig:city_layout_example}, we show prompts for PCG plugins with different functionalities, along with the corresponding API calls for each PCG plugin. We also provide a tested example for each PCG component.
These images illustrate that our systematized prompt framework enables the successful execution of different tasks. Furthermore, the adoption of a unified prompt template for all agents simplifies maintenance and facilitates consistent invocation.

\begin{algorithm}[tb]
\caption{The workflow of Scene$\mathcal{X}$}
\label{alg:algorithm}
\textbf{Input}: User query $q$, knowledge documentation $D$, The retrieved $API$ in the API collection, the retrieved $Asset_i$ in the asset collection $\Gamma$, $P_i$ represents the Prompt of each Agent, $r_1$ represents the relationshop of each asset, $End\_flag = False$. \\
\textbf{Output}: Modified API $\alpha^*$
\begin{algorithmic}[1] %[1] enables line numbers
\STATE When receiving a user query, \mymethod{} framework starts the processing workflow;
\WHILE{$End\_flag$ == False}
    \STATE Break down the input $q$ into a rough plan.
    \STATE $\{o_1, \dots, o_n\} \leftarrow LLM_{decomposition}(q, P_{decomposition})$;
    \FOR{$o_i$ \textbf{in} $\{o_1, \ldots, o_n\}$}
        \STATE $API_i$ = $Rretrieval(o_i)$.
        \IF{$API_i$ exists}
            \STATE $\alpha^* \leftarrow \text{HyperparamGen}(\alpha, P_\alpha, o_i)$
            \STATE API execution by parameter $\alpha^*$.
        \ELSE
            \STATE $Asset_i$ = $Rretrieval(o_i)$.
            \STATE Directly import $Asset_i$ into the scene.
        \ENDIF
    \ENDFOR
    \STATE $\{r_1, \dots, r_n\} \leftarrow LLM_{relationship}(o_1, \ldots, o_n)$;
    \FOR{$r_i$ \textbf{in} $\{r_1, \ldots, r_n\}$}
        \STATE $\alpha^* \leftarrow \text{HyperparamGen}(\alpha, P_\alpha, r_i)$
        \STATE API execution by parameter $\alpha^*$.
    \ENDFOR
    \IF{All the task is completed}
        \STATE $End\_flag = True$
    \ENDIF
\ENDWHILE
\end{algorithmic}
\end{algorithm}

\subsubsection{B.4. More Visualization Results}
In Fig \ref{fig:text2image}, we present the extended capabilities of \mymethod{} in urban and natural scene generation. The results demonstrate that our proposed \mymethod{} framework is capable of generating highly realistic scenes in both natural and urban environments. Furthermore, the generated content accurately corresponds to the provided texture descriptions. These outcomes serve as compelling evidence of the effectiveness and efficiency of our proposed LLM-based automatic 3D scene generation framework.
In Fig \ref{fig:No_material}, we show scenes generated by \mymethod{} without the utilization of any materials, highlighting the partial PCG components employed in the scene generation process, particularly the placement functionality achieved through geometric nodes. The results indicate that the generation of complex scenes requires the use of several sophisticated PCG plugins. Moreover, the generated scenes are solely based on object meshes, rather than relying on materials or rendering techniques. This implies that the scenes generated by \mymethod{} possess depth and other essential information, which holds significant implications across various domains.
\subsubsection{B.5. Questionnaire Interface}
In our experimental process, we employ manual evaluation to assess the results. Fig. \ref{fig:as_survey_2} shows the evaluation interface. To ensure fairness and objectivity, all evaluation images were anonymized, preventing any bias in the assessment. This approach maintained the integrity of the evaluation process, guaranteeing that each image was judged solely on its merits.

\begin{figure*}[!tbp]
\centering
\includegraphics[width=\textwidth]{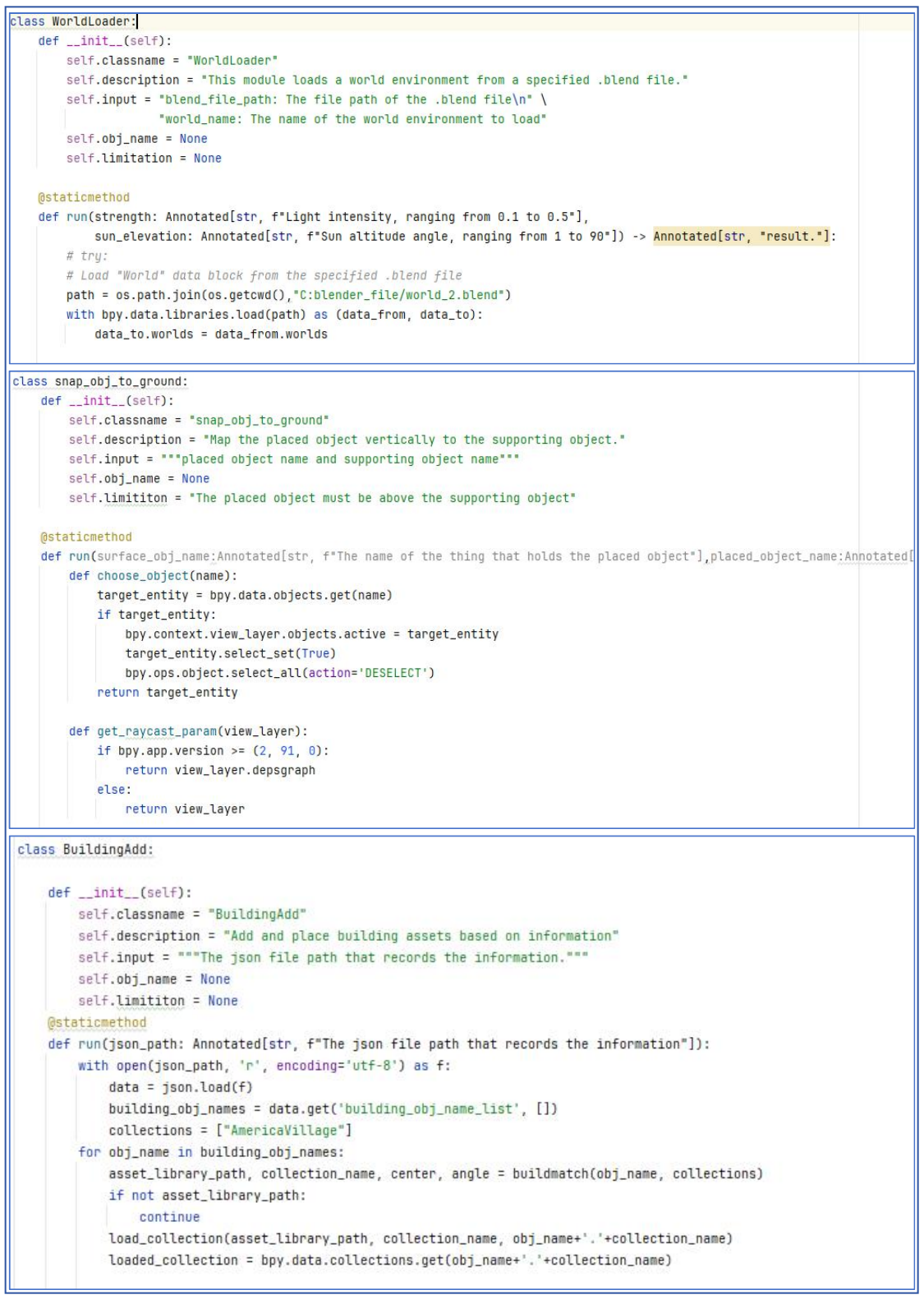}
\vspace{-0.73cm}
\caption{Standardized python template for PCG module documentation.}
\label{fig:code}
\vspace{-0.3cm}
\end{figure*}

% %planner案例
% \begin{figure*}[!tbp]
% \centering
% \includegraphics[width=\textwidth]{figs/planner_example_1.pdf}
% %\vspace{-1cm}
% \caption{Prompt Example for Planner: A Template, Code Snippet, and Specific Example.}
% \label{fig:planner}
% \vspace{-0.3cm}
% \end{figure*}

%地形案例
\begin{figure*}[!tbp]
\centering
\includegraphics[width=\textwidth]{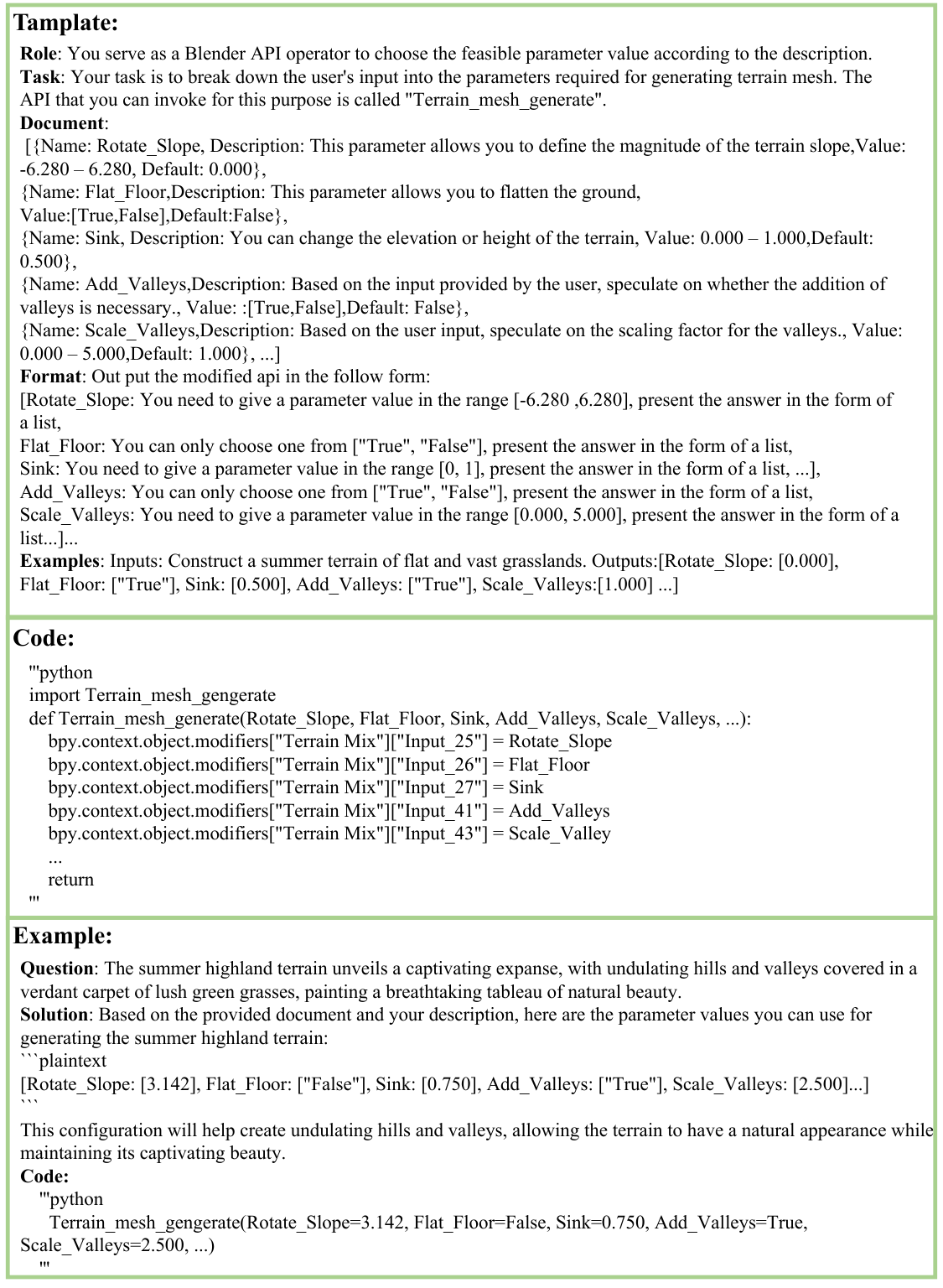}
%\vspace{-1cm}
\caption{Prompt example for terrain PCG mesh generation: a template, code snippet, and specific example.}
\label{fig:terrain_example}
\vspace{-0.3cm}
\end{figure*}

%摆放案例
\begin{figure*}[!tbp]
\centering
\includegraphics[width=\textwidth]{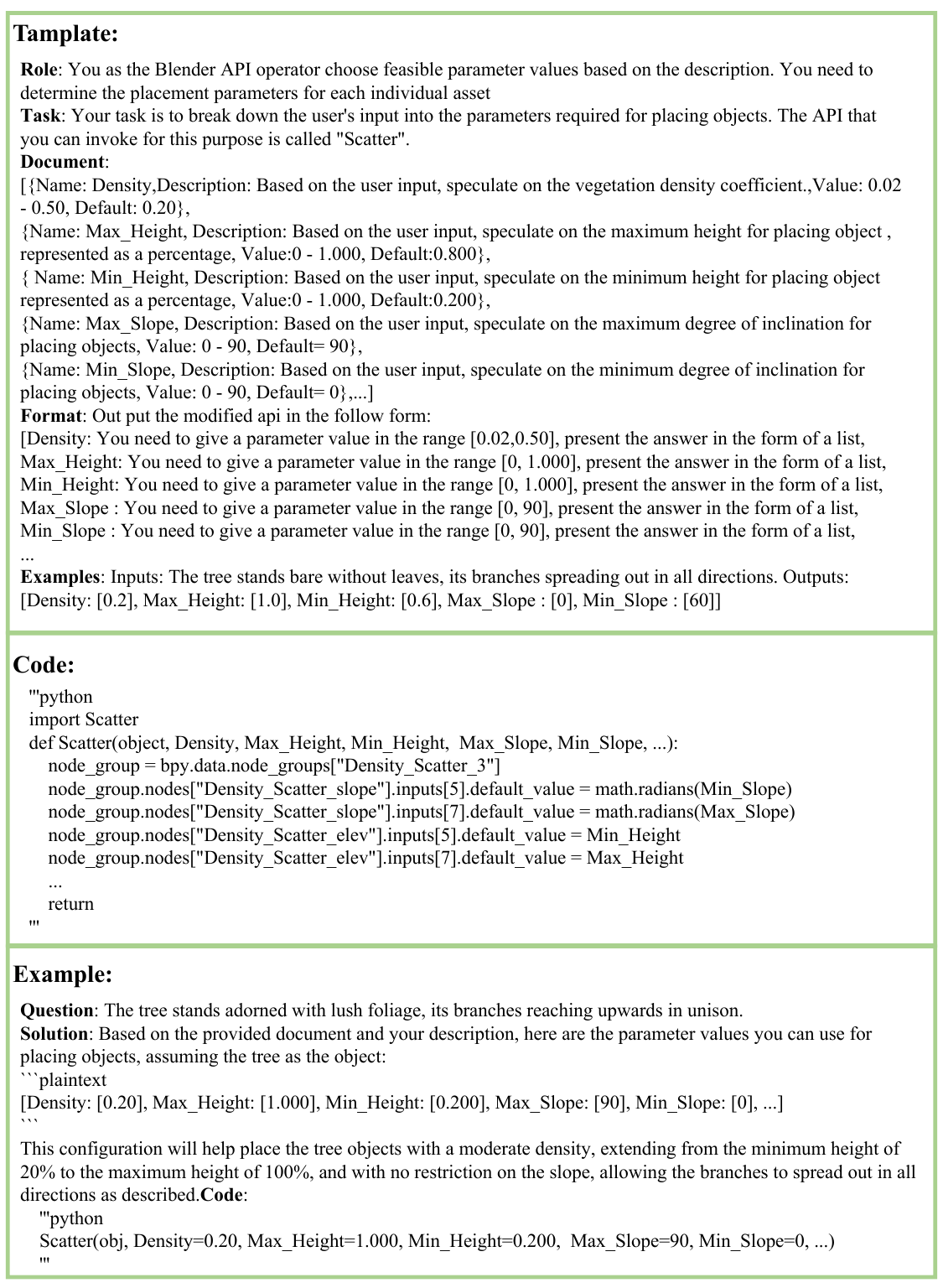}
%\vspace{-1cm}
\caption{Prompt example for object placement: a template, code snippet, and specific example.}
\label{fig:placement_example}
\vspace{-0.3cm}
\end{figure*}

%树案例
\begin{figure*}[!tbp]
\centering
\includegraphics[width=\textwidth]{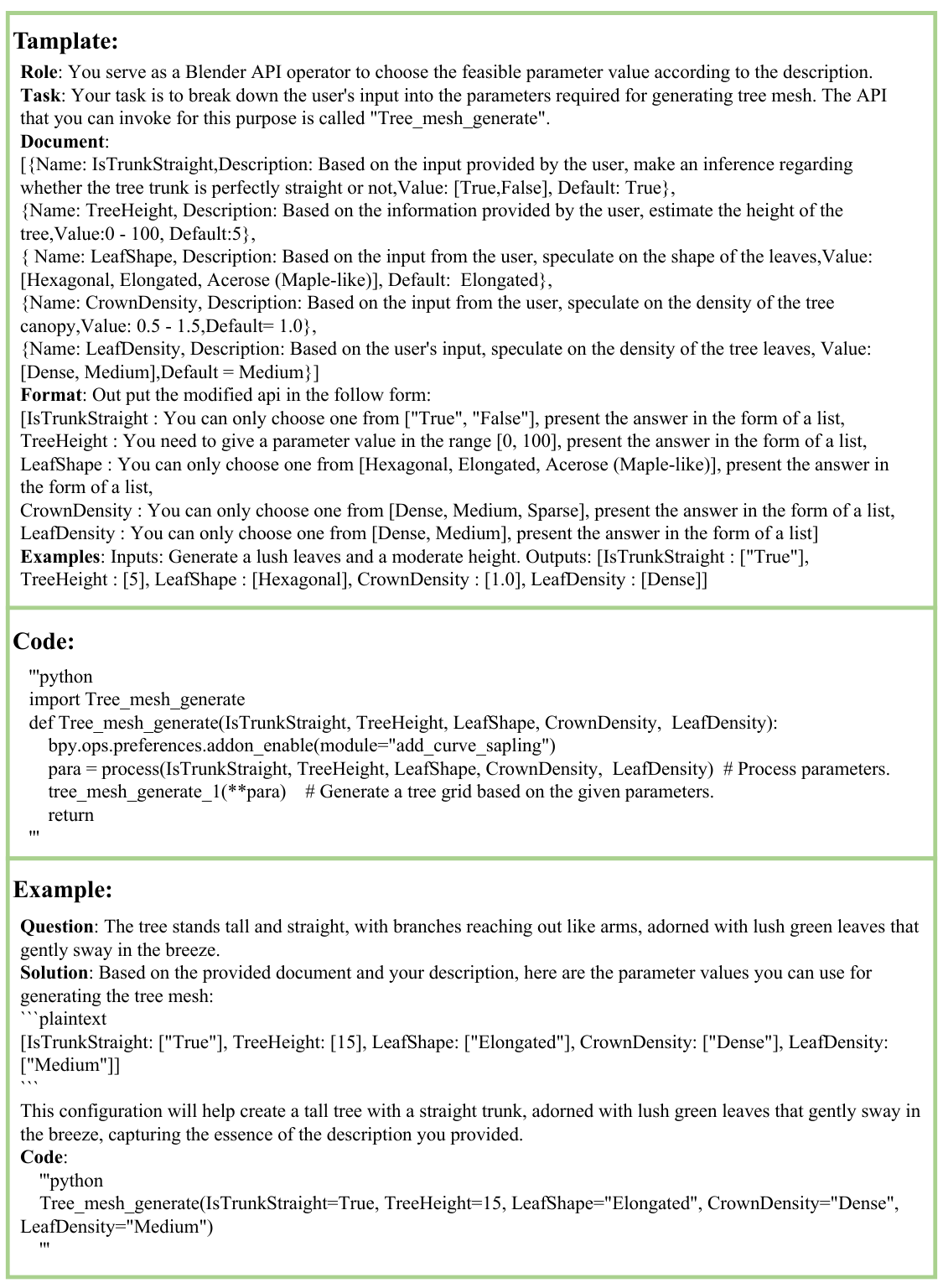}
%\vspace{-1cm}
\caption{Prompt example for tree PCG mesh generation: a template, code snippet, and specific example.}
\label{fig:tree_example}
\vspace{-0.3cm}
\end{figure*}

%city_layout案例
\begin{figure*}[!tbp]
\centering
\includegraphics[width=\textwidth]{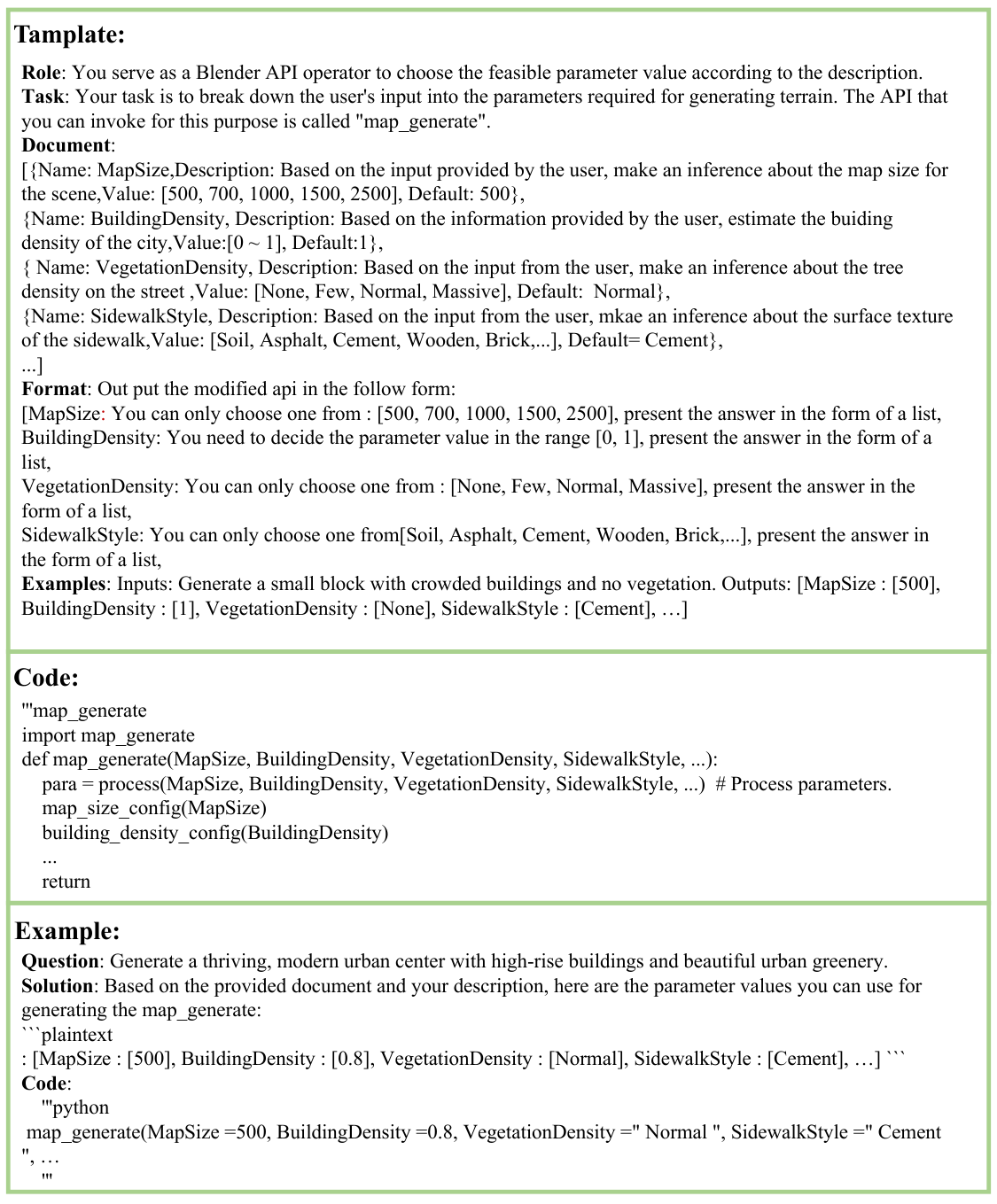}
%\vspace{-1cm}
\caption{Prompt example for city layout generation: a template, code snippet, and specific example.}
\label{fig:city_layout_example}
\vspace{-0.3cm}
\end{figure*}

%text2image
\begin{figure*}[!tbp]
\centering
\includegraphics[width=\textwidth]{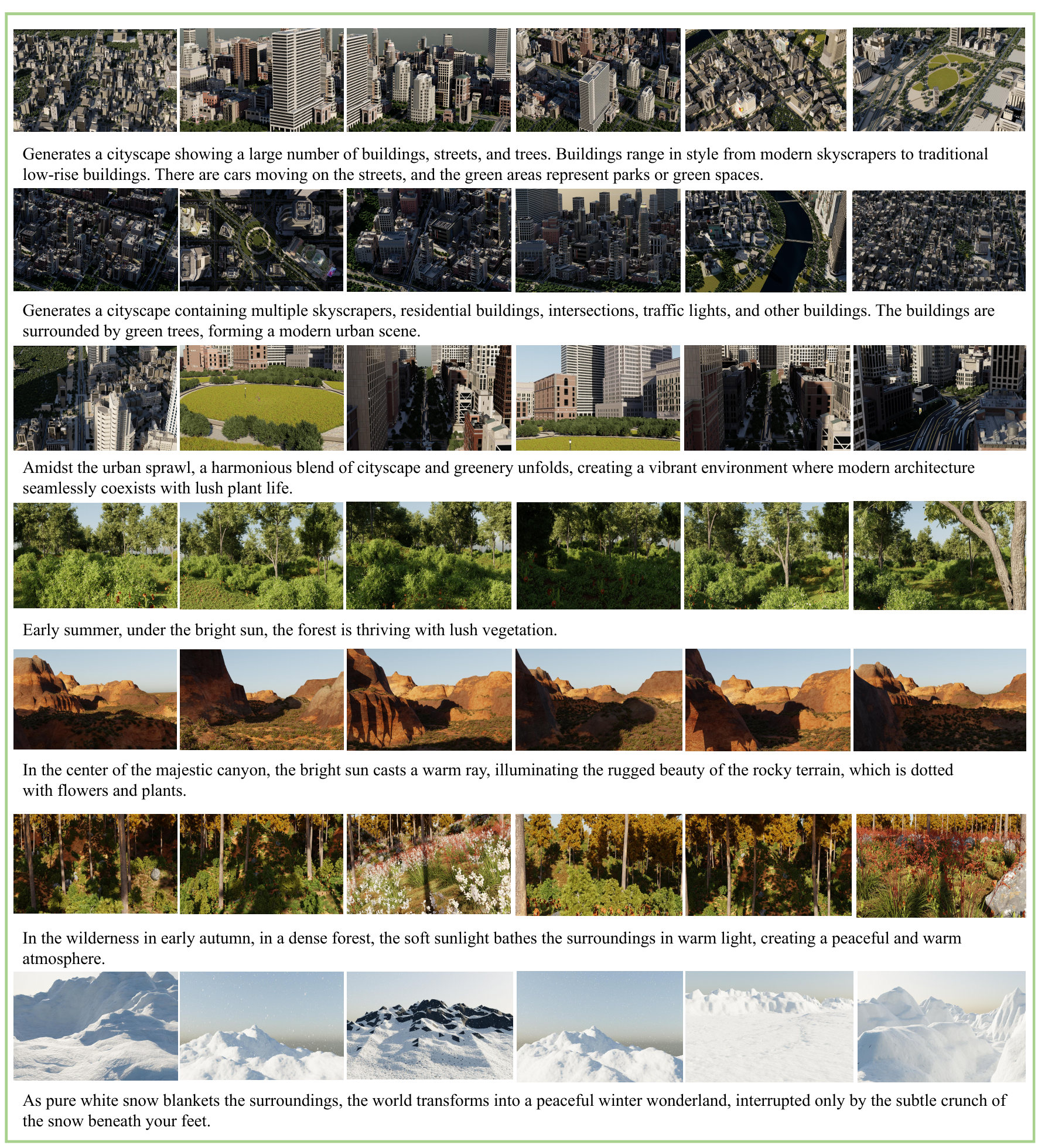}
%\vspace{-1cm}
\caption{Rendered scene results obtained by inputting text into \mymethod{}: including three urban and four natural scenes with varied prompts to generate diverse 3D environments.}
\label{fig:text2image}
\vspace{-0.3cm}
\end{figure*}

%3Dshow
\begin{figure*}[!tbp]
\centering
\includegraphics[width=\textwidth]{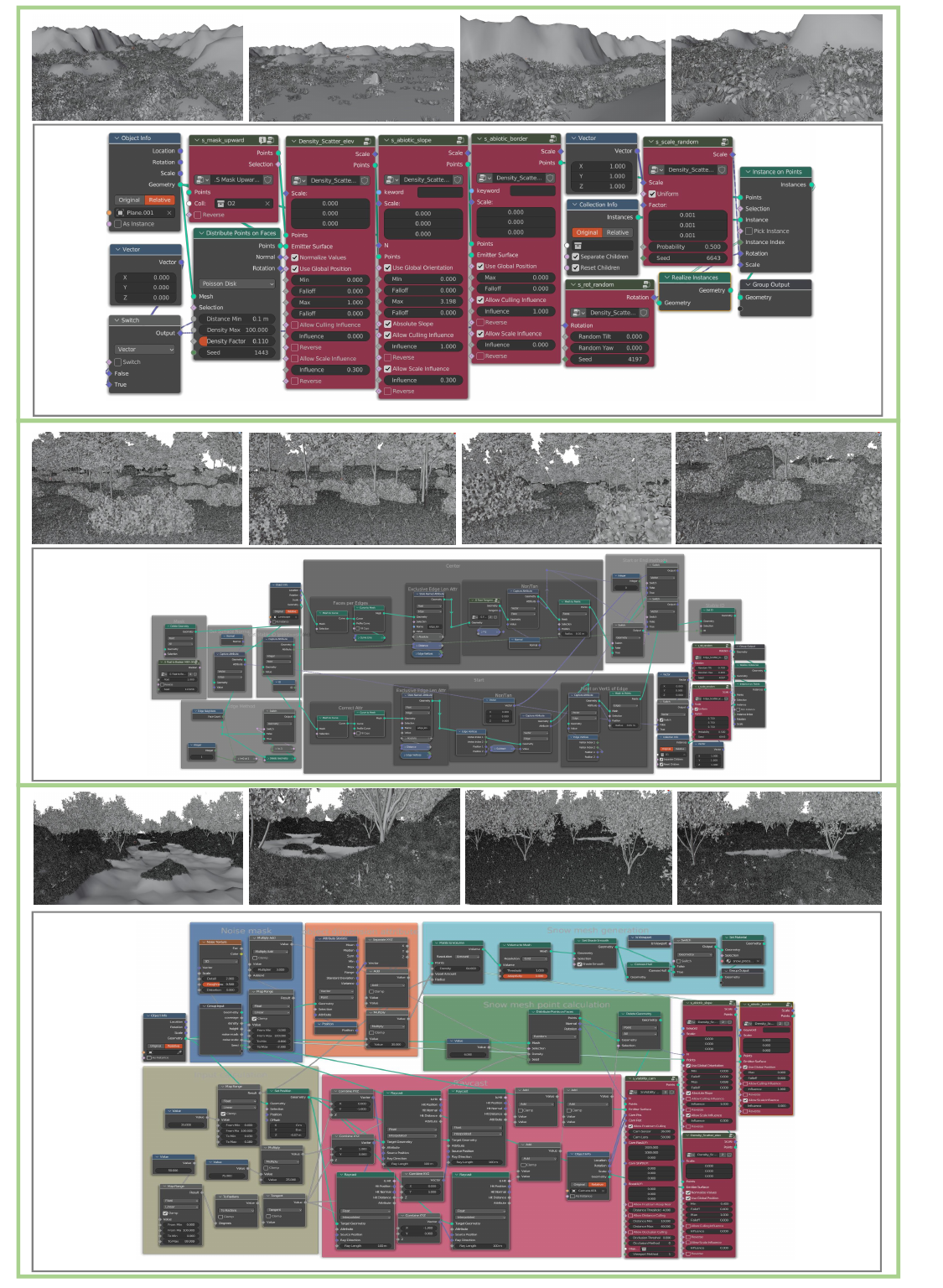}
%\vspace{-1cm}
\caption{The image illustrates scenes rendered without applying any materials and displays partial PCG components used for each scene (presented as a composition of geometry nodes).}
\label{fig:No_material}
\vspace{-0.3cm}
\end{figure*}

\begin{figure*}[!tbp]
\centering
\includegraphics[width=\textwidth]{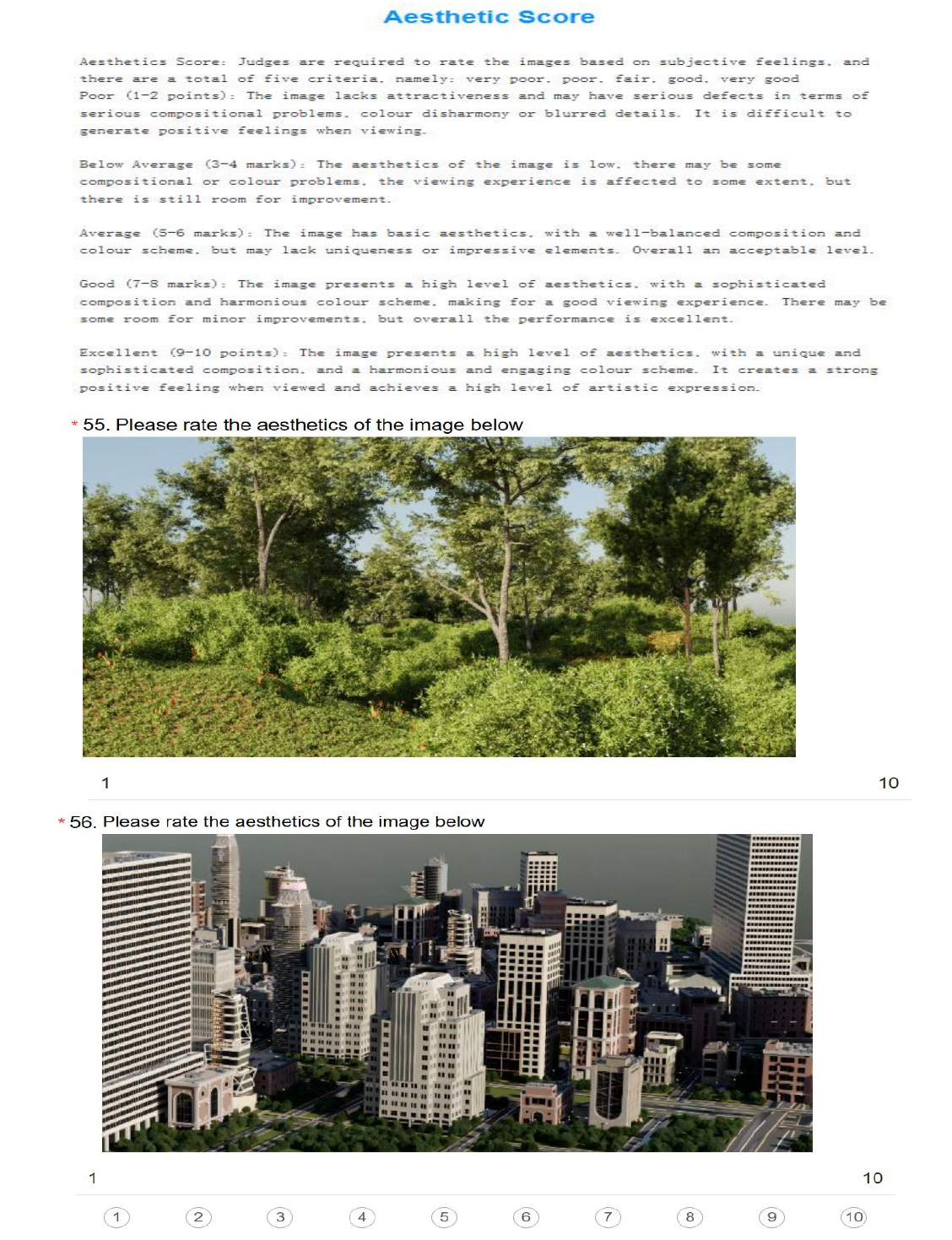}
\vspace{-0.73cm}
\caption{The image illustrates scenes rendered without applying any materials and displays partial PCG components used for each scene (presented as a composition of geometry nodes).}
\label{fig:as_survey_2}
\vspace{-0.3cm}
\end{figure*}

%\bibliographystyle{splncs04}
%\bibliography{main}
\end{document}